\documentclass{article}

\usepackage{times}
\usepackage{graphicx} 
\usepackage{subfigure} 
\usepackage{makeidx}  
\usepackage{booktabs}
\usepackage{url}
\usepackage{sidecap}
\sidecaptionvpos{figure}{c}
\usepackage{enumitem}

\usepackage{amsmath}
\usepackage{amssymb}
\usepackage{mathtools}

\usepackage{natbib}

\usepackage{algorithm}
\usepackage{algorithmic}

\renewcommand{\algorithmicrequire}{\textbf{Input:}}
\renewcommand{\algorithmicensure}{\textbf{Output:}}

\usepackage{tabularx}
\usepackage{multirow}
\usepackage{multicol}
\usepackage{pbox}
\usepackage{array}
\newcolumntype{L}[1]{>{\raggedright\let\newline\\\arraybackslash\hspace{0pt}}m{#1}}
\newcolumntype{C}[1]{>{\centering\let\newline\\\arraybackslash\hspace{0pt}}m{#1}}
\newcolumntype{R}[1]{>{\raggedleft\let\newline\\\arraybackslash\hspace{0pt}}m{#1}}

\usepackage[T1]{fontenc}
\usepackage{xcolor}
\usepackage{comment}

\def\alphabf{\boldsymbol{\alpha}}
\def\betabf{\boldsymbol{\beta}}
\def\pibf{\boldsymbol{\pi}}
\def\thetabf{\boldsymbol{\theta}}

\def\phibf{\boldsymbol{\phi}}

\usepackage[accepted]{icml2015}

\icmltitlerunning{Incremental Variational Inference for Latent Dirichlet Allocation}

\begin{document} 

\twocolumn[
\icmltitle{Incremental Variational Inference for Latent Dirichlet Allocation}

\icmlauthor{C\'{e}dric Archambeau}{cedrica@amazon.de}
\icmladdress{Amazon Development Centre Germany}
\icmlauthor{Beyza Ermi\c{s}}{ermisbeyza@gmail.com}
\icmladdress{Bogazici  University}

\icmlkeywords{stochastic incremental variational inference, incremental variational inference, asynchronous distributed computation, latent dirichlet allocation}

\vskip 0.3in
]

\begin{abstract} 
We introduce incremental variational inference and apply it to latent Dirichlet allocation (LDA). Incremental variational inference is inspired by incremental EM and provides an alternative to stochastic variational inference. Incremental LDA can process massive document collections, does not require to set a learning rate, converges faster to a local optimum of the variational bound and enjoys the attractive property of monotonically increasing it. We study the performance of incremental LDA on large benchmark data sets. We further introduce a stochastic approximation of incremental variational inference which extends to the asynchronous distributed setting. The resulting distributed algorithm achieves comparable performance as single host incremental variational inference, but with a significant speed-up.
\end{abstract} 


\section{Introduction}
\label{sec:intro}

Approximate Bayesian inference has become mainstream in machine learning~\cite{bishop2006pattern,murphy2012machine} and enjoyed a (re)gained interest in the statistics community~\cite{wang06,dunson11}. It constitutes an appealing alternative to Markov Chain Monte Carlo when one is interested in probabilistic data modelling. Approximate inference techniques are pragmatic, postulating an approximate model family and trying to find the best model within this family by optimizing a surrogate objective~\cite{wainwright08}. They are also practical, as the code implementing these inference algorithms is relatively easy to de-bug. For example, variational inference monotonically increases the variational objective. Hence, the bound provides a sanity check for correctness and can be used to monitor convergence.

The amount of data being generated and collected today is tremendous. For example, at the time of writing, there are almost 5 million articles in Wikipedia. Amazon S3 holds trillions of objects and over 6 billion hours of video are watched each month on YouTube. In 2012, the number of active Facebook users had surpassed 1 billion. The trend of ``big data growth" presents enormous challenges for industry and creates a need to invent new algorithms capable of ingesting and processing massive data sets.

Stochastic variational inference~\cite{hoffman2013stochastic} was a first step in this direction in the context of approximate inference. It relies on stochastic optimization~\cite{robbins1951stochastic} and was designed to handle very large data sets by processing the data sequentially. The drawback of stochastic variational inference requires to adjust additional parameters like the learning rate and the mini-batch size. Moreover, it does not share the attractive property of batch variational inference of monotonically increasing the bound while inferring the model parameters. 

The increasing availability of distributed architectures, such as multi-processor and grid-computing hardware, provides an opportunity to device distributed inference algorithms able to take advantage of the infrastructure and perform well at scale. Recent attempts in this direction include the work  by ~\citet{smola2010architecture,newman2009distributed,smyth2009asynchronous}. However, stochastic variational inference cannot easily be adapted to the distributed optimization setting.

To address the shortcomings of stochastic variational inference, we introduce incremental variational inference, which generalizes incremental EM proposed by \citet{neal1998view}.  Like stochastic variational inference, incremental variational inference processes the data sequentially. However, it does not require to adjust the learning rate. By maintaining a set of local statistics, it also preserves the property of monotonically increasing the variational objective at each iteration. We further propose a stochastic modification of incremental variational inference that can be executed in a distributed environment. We report significant horizontal speed-up, while sacrificing very little predictive performance.

In this paper, we focus on Latent Dirichlet Allocation (LDA)~\cite{griffiths2004finding, blei2003latent}, a popular generative model for documents. However, it should be noted that the approximate inference scheme we introduce is general and it is applicable to any latent variable model with a set of local and global variables.

Topic Models like LDA make the simplifying assumption that documents can be represented as bag-of-words. This means they ignore the sequential structure of the text. More specifically, LDA postulates the existence of a collection of $K$ topics, each of which is defined as a categorical distribution over a vocabulary of size $V$. It further assumes that each document in a corpus of $D$ documents is generated according to a document-specific categorical distribution over these topics.

Let us denote word $n$ in document $d$ by $x_{nd}$ and its topic assignment by $z_{nd}$. The generative model is defined as follows:
\begin{align}
z_{nd} \mid \thetabf_d 
	&\sim \text{Categorical} (\thetabf_d)  ,\notag\\
 x_{nd} \mid z_{nd} \text{,} \lbrace \phibf_k \rbrace_{k=1}^K 
 	&\sim \text{Categorical} (\phibf_{z_{nd}})  ,
\end{align}
where $\thetabf_d\sim\text{Dirichlet} (\alpha_0 \textbf{1}_{K})$ and $\phibf_k \sim \text{Dirichlet} (\beta_0 \textbf{1}_{V})$. The parameters $\alpha_0$ and $\beta_0$ are non-negative reals.

The paper is organized as follows. In Section~\ref{sec:mvi}, we review batch and stochastic variational inference for LDA. In Section~\ref{sec:ivi}, we introduce incremental variational inference and its stochastic counterpart. The asynchronous distributed inference algorithm for LDA is described in Section~\ref{sec:adivi}. After discussing related work in Section~\ref{sec:related}, we present results on several large benchmark data sets in Section~\ref{sec:exp_res}.


\section{Variational Inference for LDA}
\label{sec:mvi}

Bayesian inference is often difficult in practice as it requires the computation of analytically intractable integrals. One can circumvent this problem by resorting to Markov Chain Monte Carlo (MCMC) to simulate samples from the posterior. For example, collapsed Gibbs sampling has proven to be very successful for inference in LDA~\cite{griffiths2004finding}. 
However, convergence of MCMC is notoriously difficult to verify. A more pragmatic approach is to consider deterministic approximations like variational inference \cite{bishop2006pattern} or expectation propagation \cite{minka2002expectation}. These methods turn the inference problem into an optimization problem, which is often more easy to tackle and to monitor convergence.

Variational inference maximizes a lower bound to the log marginal likelihood of the data by approximating the true posterior by postulating a simpler distribution, which is parametrized by a set of free parameters. In the case of LDA, the variational bound is given by
\begin{align*}
\ln p(X) &\geqslant \langle\ln p(X,Z,\Theta,\Phi) \rangle + \text{H} [q(Z,\Theta,\Phi)] \notag \\
             & = \ln p(X) - \text{KL} [q(Z,\Theta,\Phi) \| p(Z,\Theta,\Phi |X)] ,
\end{align*} 
where $X = \lbrace x_{nd}\rbrace_{n,d} $, $Z = \lbrace z_{nd}\rbrace_{n,d} $, $\Theta = \lbrace \thetabf_{d}\rbrace_{d} $ and $\Phi = \lbrace \phibf_{k}\rbrace_{k} $. The notation $\langle\cdot\rangle$ denotes an expectation wrt $q(Z,\Theta,\Phi)$, $\text{H} [p]$ is the differential entropy and $\text{KL} [q \| p]$ is the Kullback-Leibler divergence wrt $q$. Maximizing this bound is equivalent to minimising the Kullback-Leibler divergence between the true posterior $p(Z,\Theta,\Phi |X)$ and the approximate posterior $q(Z,\Theta,\Phi)$. In general, this minimization problem is still problematic, unless we further restrict the form of $q(Z,\Theta,\Phi)$.

Mean field variational inference (MVI) assumes the latent variables and the parameters are independent when conditioning on the data, that is, $q(Z,\Theta,\Phi) = \prod_{n,d}q(z_{nd}) \times \prod_d q(\thetabf_d) \times \prod_kq(\phibf_k)$. It is easy to show that in this case the lower bound is maximised when the factors are defined as follows~\cite{blei2003latent}:
\begin{align}
q(z_{nd})
	&= \text{Categorical}(\pibf_{nd}) ,
&\pi_{knd}
	&\propto e^{\langle \ln\theta_{kd} \rangle + \langle \ln\phi_{x _{nd}k} \rangle} ,   \notag  \\
q(\thetabf_{d})
	&= \text{Dirichlet}(\alphabf_{d}) ,
&\alpha_{kd}
	&= \alpha_0 + \langle m_{kd} \rangle ,\notag   \\
q(\phibf_{k})
	&= \text{Dirichlet}(\betabf_{k}) ,
&\beta_{vk}
	&= \beta_0 + \langle m_{vk} \rangle ,\label{eq:mvi}
\end{align}
where $m_{kd}$ is the (unobserved) number of times topic $k$ appeared in document $d$ and $m_{vk}$ the (unobserved) number of times word token $v$ was assigned to topic $k$ in the corpus. Hence, the special quantities $\langle m_{kd} \rangle$ and $\langle m_{vk} \rangle$ are expected counts under the variational approximation. They are respectively given by $\sum_n \pi_{knd} $ and $\sum_{n,d} \delta_v (x_{nd}) \pi_{knd}$. The function $\delta_v(\cdot)$ is Dirac's delta centred at $v$. The expectations $\langle \ln \theta_{kd} \rangle$ and $\langle \ln \phi_{vk} \rangle$ are respectively given by $\psi(\alpha_{kd}) - \psi(\sum_k \alpha_{kd})$ and $\psi(\beta_{vk}) - \psi(\sum_v \beta_{vk})$. 

MVI is a coordinate ascent method that converges to a local maximum of the variational bound~\cite{beal2003variational}. Cycling through the updates for variational parameters in (\ref{eq:mvi}) ensures a monotonic increase of this bound. MVI is a batch inference approach: every update of the variational  parameter $\beta_{vk}$ requires updating all word-specific proportions $\pibf_{nd}$ beforehand, which is costly when the corpus is large. Stochastic variational inference (SVI) was recently proposed in the context of LDA to address this problem~\cite{hoffman2010online,hoffman2013stochastic}. The goal was to speed up inference and to scale up LDA to very large data sets.

SVI optimizes the lower bound by stochastic optimization~\cite{robbins1951stochastic}. It maintains a set of local and global parameters, which characterize the variational posteriors. Local variables are the indicator variables $Z$ and the document-topic proportions $\Theta$, which are respectively characterized by the local parameters $\{\pibf_{nd}\}_{n,d}$ and $\{\alpha_d\}_d$. The global variables are topic-word proportions $\Phi$, which are characterized by the global parameters $\{\betabf_k\}_k$. SVI considers a noisy, but unbiased estimate of the gradients of the variational parameters associated to the global variables.

This leads to the following updates (document d being picked at random)~\cite{hoffman2013stochastic}:
\begin{align}
\betabf_k^{(t)} &= (1-\rho_t) \betabf_k^{(t-1)} + \rho_t \hat{\betabf}_k ,  \notag \\
\hat{\beta}_{vk} &= \beta_0 + D \sum_{n=1}^{N_d} \delta_v (x_{nd}) \pi_{knd} ,\label{eq:svi}
\end{align} 
where $\sum_t \rho_t = \infty$ and $\sum_t \rho_t^{2} < \infty$. Throughout this work, we will use the learning rate $\rho_t=(t+\tau)^{-\kappa}$, where $\kappa\in(0.5,1]$ and $\tau\geqslant 0$.

Intuitively, the second term on the right hand side of (\ref{eq:svi}) is a noisy, but unbiased estimate of the expected number of counts appearing in (\ref{eq:mvi}), namely $\langle m_{vk} \rangle$. The variational parameters associated to the local variables (that is, $\pibf_{nd}$ and $\alphabf_d$) can be computed as in MVI. Typically, mini-batches are used to stabilize the gradients. An interesting property of SVI is that it corresponds to natural gradients with respect to the variational distribution~\cite{hoffman2010online}. 

The intrinsic noise of the stochastic gradients can impede the convergence of SVI. Variance reduction techniques have been proposed to address this issue \cite{wang2013variance,paisley2012variational}. SVI is also sensitive to the learning rate decay schedule and choice of mini-batch size~\cite{ranganath2013adaptive}. Next, we derive incremental variational inference for LDA, which does not require to choose and adjust the learning rate. Importantly, it ensures a monotonic increase of the bound and convergence to a local maximum of the log marginal likelihood like MVI.

\section{Incremental Variational Inference for LDA}
\label{sec:ivi}

Incremental variational inference (IVI) computes updates in a similar fashion as incremental EM~\cite{neal1998view}. Each iteration performs a partial variational E-step before performing a variational M-step. This amounts to maintaining a set of global statistics associated to the global variables, which are updated incrementally in the variational E-step by first subtracting the old statistics associated to a data point (or a mini-batch) and adding back the corresponding new one. The updated global statistics are then used in the variational M-step. This is to be contrasted with SVI. Indeed, SVI uses a noisy estimate of the global statistics, which is based exclusively on the mini-batch that is considered in the current iteration. In the case of LDA, IVI leads to the following incremental update:
\begin{align}
\beta_{vk}  &= \beta_0 + \langle m_{vk} \rangle + \sum_{n=1}^{N_d} \delta_v (x_{nd})  \big(\pi_{knd}^{(t)} - \pi_{knd}^{(t-1)} \big) ,
\label{eqn:betaIVI}
\end{align}
while the updates for $\pibf_{nd}$ and $\alphabf_{d}$ are the same as in MVI as they are associated to the local variables. The main advantage of IVI is that it ensures a monotonic increase of the bound and does not require to have seen all the data points to make progress. The price we have to pay is that we have to store the previous set of proportions $\pibf_{nd}$, which can be costly when the number of topics $K$ is large as the additional memory requirements scale as a constant factor times the number of words in the corpus. IVI for LDA is summarized in Algorithm~\ref{alg:ivi}.

\begin{algorithm}[t]
\caption{Incremental Variational Inference (IVI)}
\begin{algorithmic}[1]
\STATE  Initialize $\beta_{vk}^{(0)}$ randomly; set $\alpha_{kd}$ = $\alpha_0$.
\FOR{$t = 1, 2,\cdots$}
	\STATE Sample a document $d$ uniformly 
		\REPEAT
			\STATE $\pi_{knd}^{(t)} \propto e^{\langle \ln\theta_{kd} \rangle + \langle \ln\phi_{x _{nd}k} \rangle}$
			\STATE $\alpha_{kd} = \alpha_0 + \sum_{n=1}^{N_d} \pi_{knd} $
		\UNTIL{$\alpha_{kd}$ and $\pi_{knd}$ converge.}	
		\STATE $\beta_{vk} = \beta_0 + \langle m_{vk} \rangle + \sum_{n=1}^{N_d} \delta_v (x_{nd})  \big(\pi_{knd}^{(t)} - \pi_{knd}^{(t-1)} \big)$
\ENDFOR
\end{algorithmic}\label{alg:ivi}
\end{algorithm}


Subsequently, we will also consider a stochastic variant of the IVI algorithm (S-IVI), which is closely related to stochastic average gradient (SAG) descent~\cite{roux2012stochastic}, which maintains a running average of the gradient. SAG has the low iteration cost of stochastic gradient descent and the linear convergence rate of batch gradient descent. S-IVI requires to set a learning rate, but it is amenable to the distributed variant discussed in the next section. It does not maintain strictly accurate sufficient statistics, rather it uses statistics computed as decaying average of recently visited data points. The resulting update is given by
\begin{align}
\betabf_k^{(t)}   &= (1-\rho_t) \betabf_k^{(t-1)} + \rho_t \hat{\betabf}_k \text{,}  \notag \\
\hat{\beta}_{vk} & = \beta_0 + \langle m_{vk} \rangle + \sum_{n=1}^{N_d} \delta_v (x_{nd})  \big(\pi_{knd}^{(t)} - \pi_{knd}^{(t-1)} \big) ,
\label{eqn:betaSIVI}
\end{align}
where $\rho_t=(t+\tau)^{-\kappa}$ as in SVI.


\section{Distributed Variational Inference for LDA}
\label{sec:adivi}

To speed up inference in the context of large data sets, SVI and IVI process document sequentially. In this section, we further scale up IVI by extending it to the distributed setting. We introduce asynchronous distributed incremental variational inference (D-IVI), which infers topics comparable to those inferred by S-IVI, but with a significant reduction in computation time.

Distributed inference algorithms handle multiple mini-batches in parallel to leverage distributed infrastructures. The key advantage of an asynchronous algorithm over a synchronous one is that it does not require a global synchronization step. As a result, it is not limited by the speed of the slowest processor (or worker). Moreover, the algorithm needs to be fault-tolerant, meaning that it needs to be robust to delays and possibly inaccurate updates. 

The S-IVI updates in Section~\ref{sec:ivi} are amenable to a distributed implementation with one master and $P$ workers, each of which holds $1/P$ of the documents in the corpus. The workers hold the local parameters $\{\pibf_{nd}\}_{n,d}$ and $\{\alphabf_d\}_d$. They independently carry-out a variational E-step based on their possibly outdated copies of the global parameters $\{\betabf_k\}_k$. Once they are done, they send the corrected statistics associated to the mini-batch to the master, that is, $\sum_{n=1}^{N_d} \delta_v (x_{nd})  \big(\pi_{knd}^{(t)} - \pi_{knd}^{(t-1)} \big)$. The masters updates the global parameters according to (\ref{eqn:betaSIVI}) and sends back the updated value to the worker. In practice, there is a trade-off between the convergence speed and the amount of communication. Smaller mini-batches speed up convergence of the algorithm, but increase the communication overhead. Algorithm~\ref{alg:divi} summarizes D-IVI.



\begin{algorithm}[t]
\caption{Distributed IVI (D-IVI)}
\begin{algorithmic}[1]
\STATE Initialize $\beta^{(0)}$ randomly; set $\alpha_{kd}$ = $\alpha_0$. 
\STATE Set the step-size schedule $\rho_t$
\STATE Split documents into $P$ disjoint subsets $\lbrace D_1, \cdots, D_P \rbrace$
\FOR{$t = 1, 2,\cdots \infty$}
	\FOR {each processor $p \in \lbrace1, \cdots, P \rbrace$ in parallel}
	\STATE Sample a document $d$ uniformly from $D_p$
		\REPEAT
			\STATE $\pi_{knd}^{(t)} \propto e^{\langle \ln\theta_{kd} \rangle + \langle \ln\phi_{x _{nd}k} \rangle}$
			\STATE $\alpha_{kd} = \alpha_0 + \sum_{n=1}^{N_d} \pi_{knd} $
		\UNTIL{$\alpha_{kd}$ and $\pi_{knd}$ converge.}
	\ENDFOR		
\STATE $\hat{\beta}_{vk} = \beta_0 + \langle m_{vk} \rangle + \sum_{n=1}^{N_d} \delta_v (x_{nd})  \big(\pi_{knd}^{(t)} - \pi_{knd}^{(t-1)} \big) $
\STATE $\betabf_k^{(t)} = (1-\rho_t) \betabf_k^{(t-1)} + \rho_t \hat{\betabf}_k $
\ENDFOR
\end{algorithmic}\label{alg:divi}
\end{algorithm}

We conclude this section by noting that the SVI updates cannot be applied in the asynchronous distributed setting. Even in the case of only two processors, we encountered numerical issues. Even when taking small step sizes, we not able to ensure the convergence of the algorithm due to the stale global parameters.


\section{Related Work}
\label{sec:related}

Collapsed variational inference for LDA \cite{teh2006collapsed} is the de facto standard for learning topic models on corpora of moderate size. Recently, SVI was introduced to scale up inference and making it possible to handle massive corpora~\cite{hoffman2010online,hoffman2013stochastic}. \citet{foulds2013stochastic} take this work one step further by developing stochastic collapsed variational inference. Modifications of SVI, such as subsampling from data non-uniformly~\cite{gopalan2013scalable} or using control variates~\cite{wang2013variance}, have been proposed to reduce the variance in the the noisy gradient and further speed up convergence. Along similar lines, \citet{paisley2012variational} develop an algorithm that allows for direct optimization of the variational lower bound for variance reduction in stochastic gradient and \citet{mandt2014smoothed} propose a variance reduction scheme tailored to SVI by averaging successively over the sufficient statistics of the local variational parameters. 
All these methods, however, require to tune at least the learning rate and the mini-batch size. By contrast, IVI uses an incremental method to reduce the variance of the noisy natural gradient and has no learning rate. Our work is most closely related to the work by \citet{hughes2013memoized}. They generalize previous incremental variants of the EM algorithm and develop the memoized online variational inference algorithm which is analogous to IVI, but they do not consider the stochastic and distributed extensions of IVI.

Various implementations and improvements have been explored for developing distributed algorithms for LDA to improve scalability in terms of memory and computation. Most works consider parallel algorithms that are synchronous. Besides, these studies parallelize batch variational inference. For example, \citet{nallapati2007parallelized} describe distributed mean-field variational EM for LDA. Like in the case of D-IVI it relies on the fact that the expensive variational E-step can easily be parallelized because the local variable are conditionally independent. However, the master node waits until each of the workers completes its job to perform the M-step. \citet{wolfe2008fully} investigate the parallelization of both the E- and M-step of variational EM for LDA. Each node computes partial statistics in a local E-Step, sends these to a central node, and receives back completed statistics relevant for completing its local M-Step. This distributed version of LDA produces identical results to the sequential version of the algorithm but it requires a global synchronization step. \citet{zhai2012mr} proposed a distributed variational inference algorithm using the MapReduce framework, where the E-step is done in the Mappers and the M-step in the Reducer. 
Another set of works attempt to distribute MCMC algorithms ~\cite{smola2010architecture, newman2009distributed, nallapati2007parallelized,thiesson2001accelerating,wolfe2008fully}, where workes concurrently run several Gibbs samplers and perform a global update of the topic counts after the synchronization. Up to our knowledge, only \citet{smyth2009asynchronous} propose an asynchronous approach for LDA, which is based on Gibbs sampling unlike D-IVI.


\section{Experiments and Results}
\label{sec:exp_res}

We carry out two types of experiments. First, we study the performance of IVI and S-IVI for LDA. We also benchmark IVI against MVI and SVI on large document collections. Second, we measure speed-ups that are obtained with our distributed algorithm (D-IVI).
\renewcommand{\arraystretch}{1.2}
\begin{table*}[ht!]
\caption{Characteristics of data sets used in experiments.}
\begin{center}
\hspace*{-3mm}
\scalebox{0.85}{
    \begin{tabular}{  l  p{15mm}  p{15mm}  p{15mm}  p{15mm}  p{27mm}  p{15mm} }
   &   AP &  Newsgroup  & Wikipedia  &   Arxiv  & Customer Review  &  NYT   \\  \hline
    Number of documents in training set        & 1246       & 13888    & 39565  &   782385  &   452944  &   290000  \\ 
    Number of documents in test set             & 1000       &  5000      & 10000  &  100000   &  100000   &   10000    \\ 
	Average number of words per document  & 198         & 249         & 260      &  116          &    151       &   232  \\ 
    Number of words in vocabulary  &  10473   &  27059  &   42419  &  141927  &  120043  &   102660   \\ 
    \end{tabular}
    \label{table:datasets}}
\end{center}
\end{table*}
\begin{figure*}[ht!]
\begin{minipage}[b]{0.24\linewidth}
\centering
\includegraphics[scale=0.23]{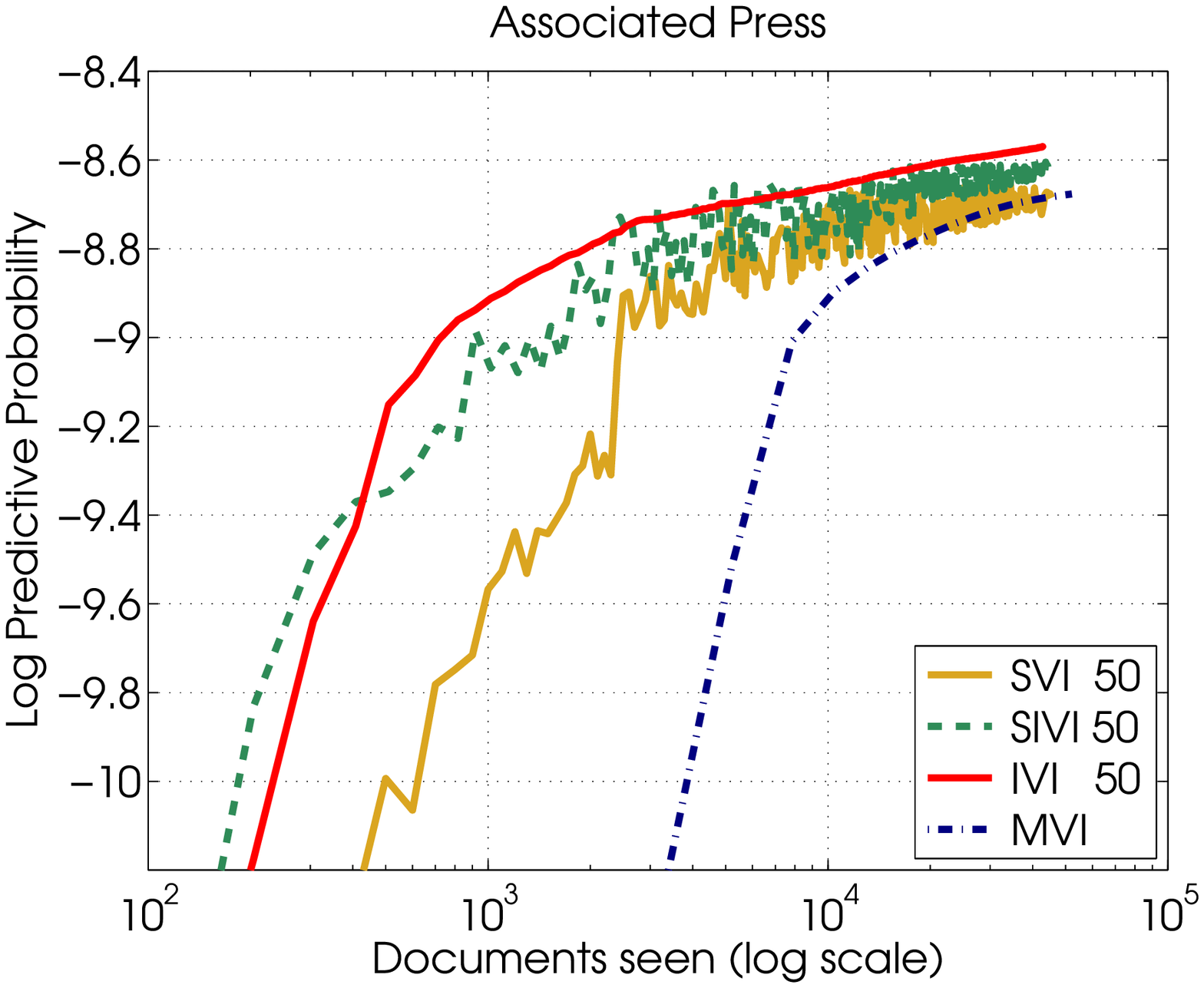}
\end{minipage}
\begin{minipage}[b]{0.24\linewidth}
\centering
\includegraphics[scale=0.23]{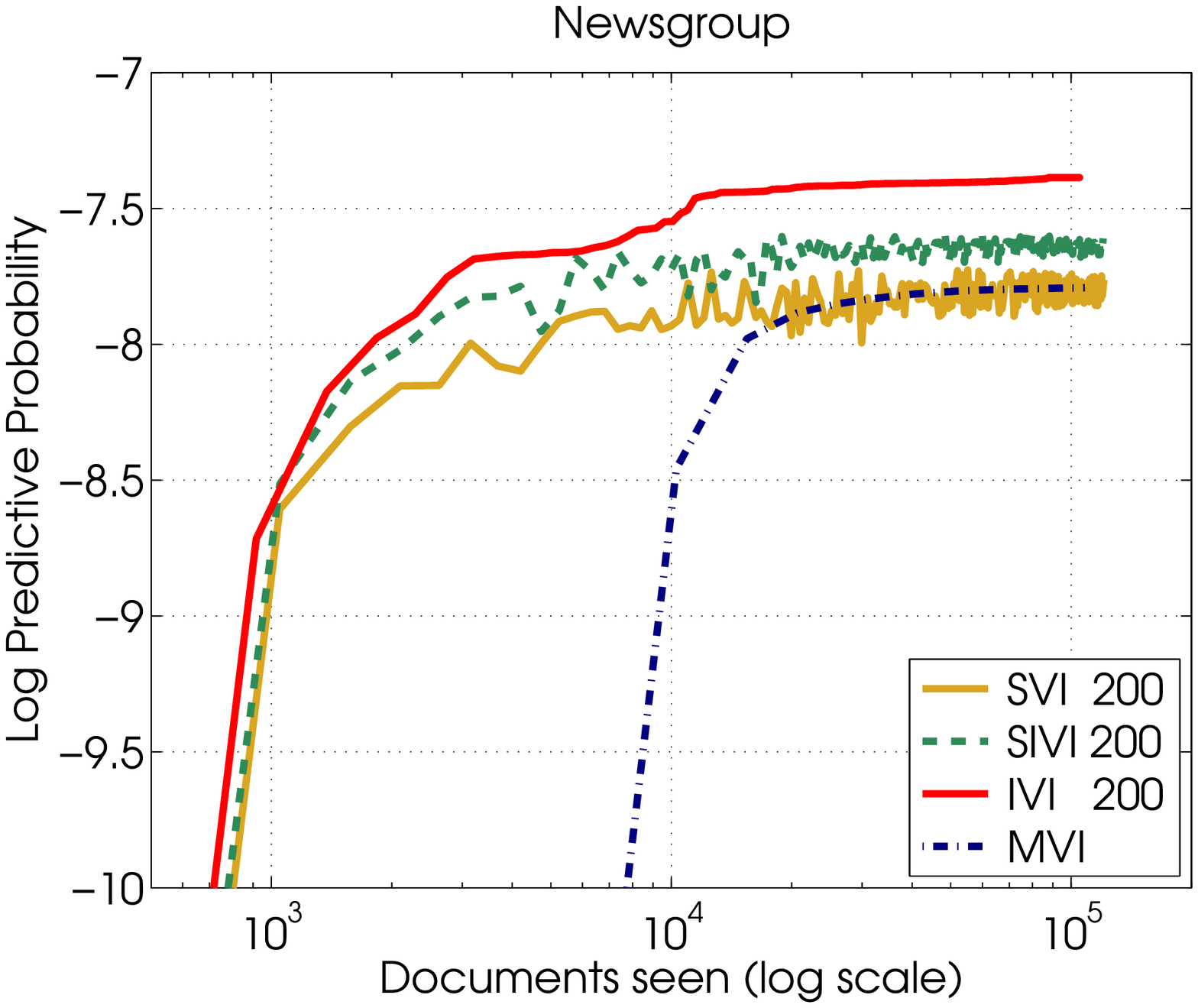}
\end{minipage}
\begin{minipage}[b]{0.24\linewidth}
\centering
\includegraphics[scale=0.23]{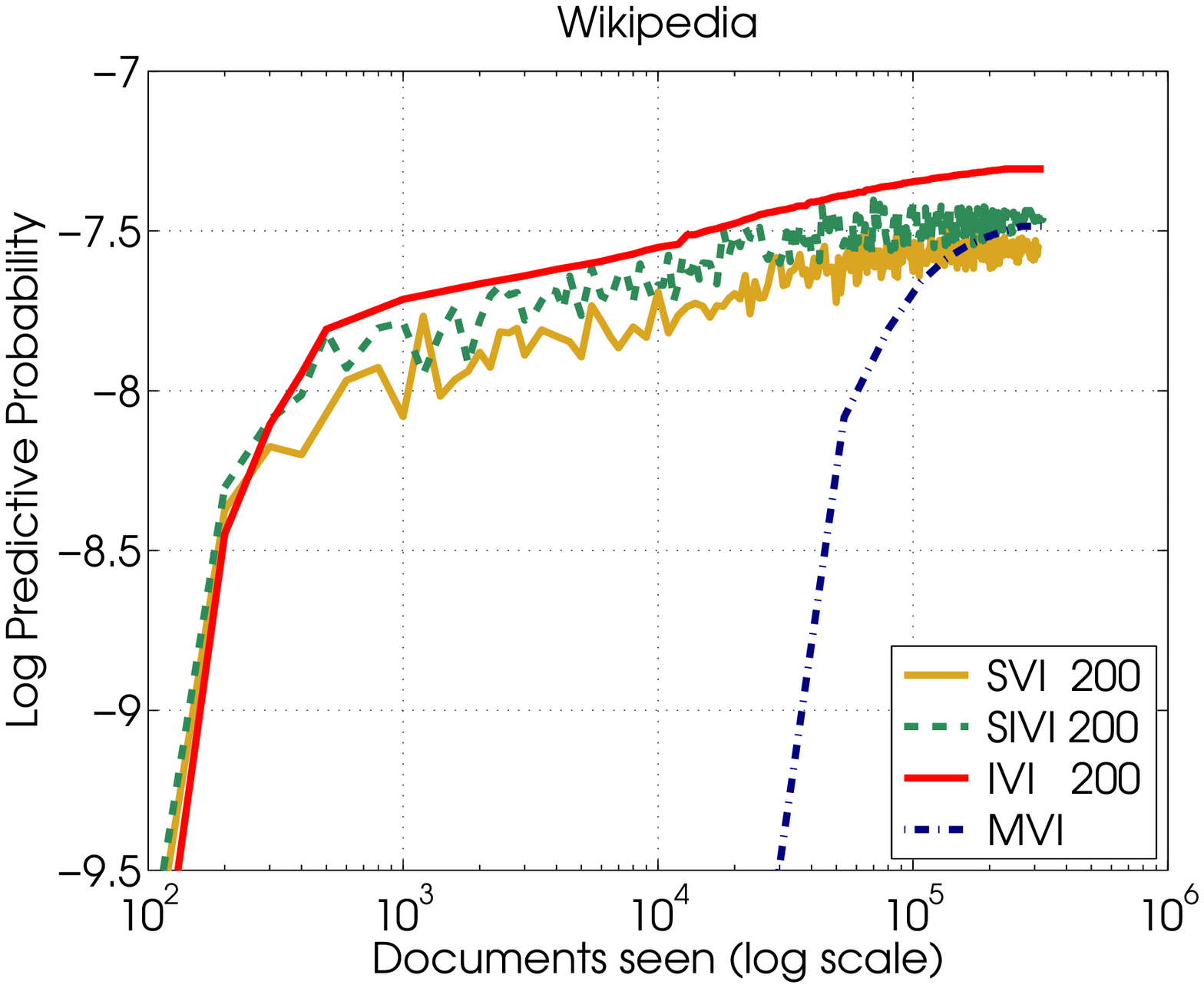} 
\end{minipage}
\begin{minipage}[b]{0.24\linewidth}
\centering
\includegraphics[scale=0.23]{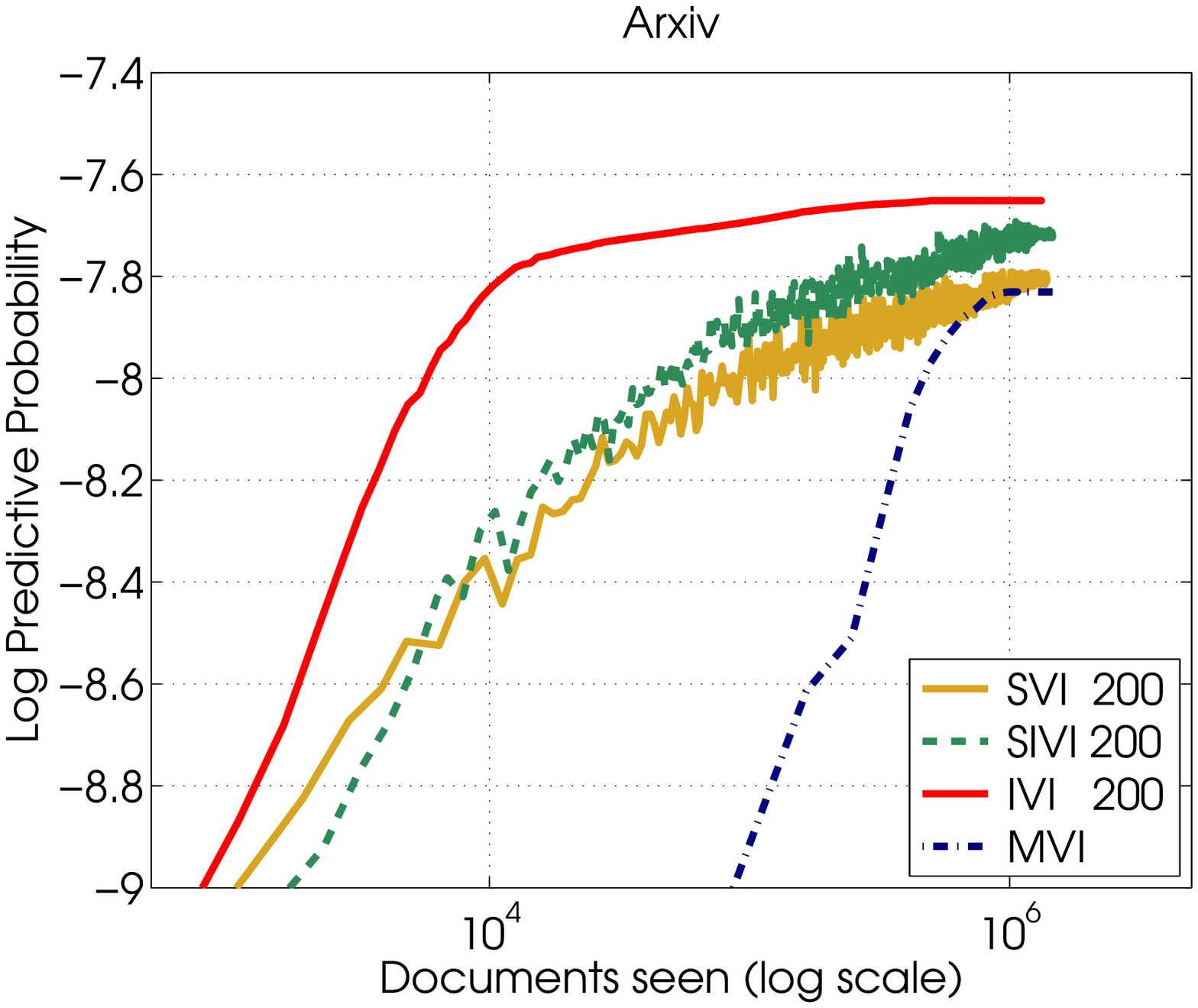} 
\end{minipage}
\\
\begin{minipage}[b]{0.24\linewidth}
\centering
\includegraphics[scale=0.23]{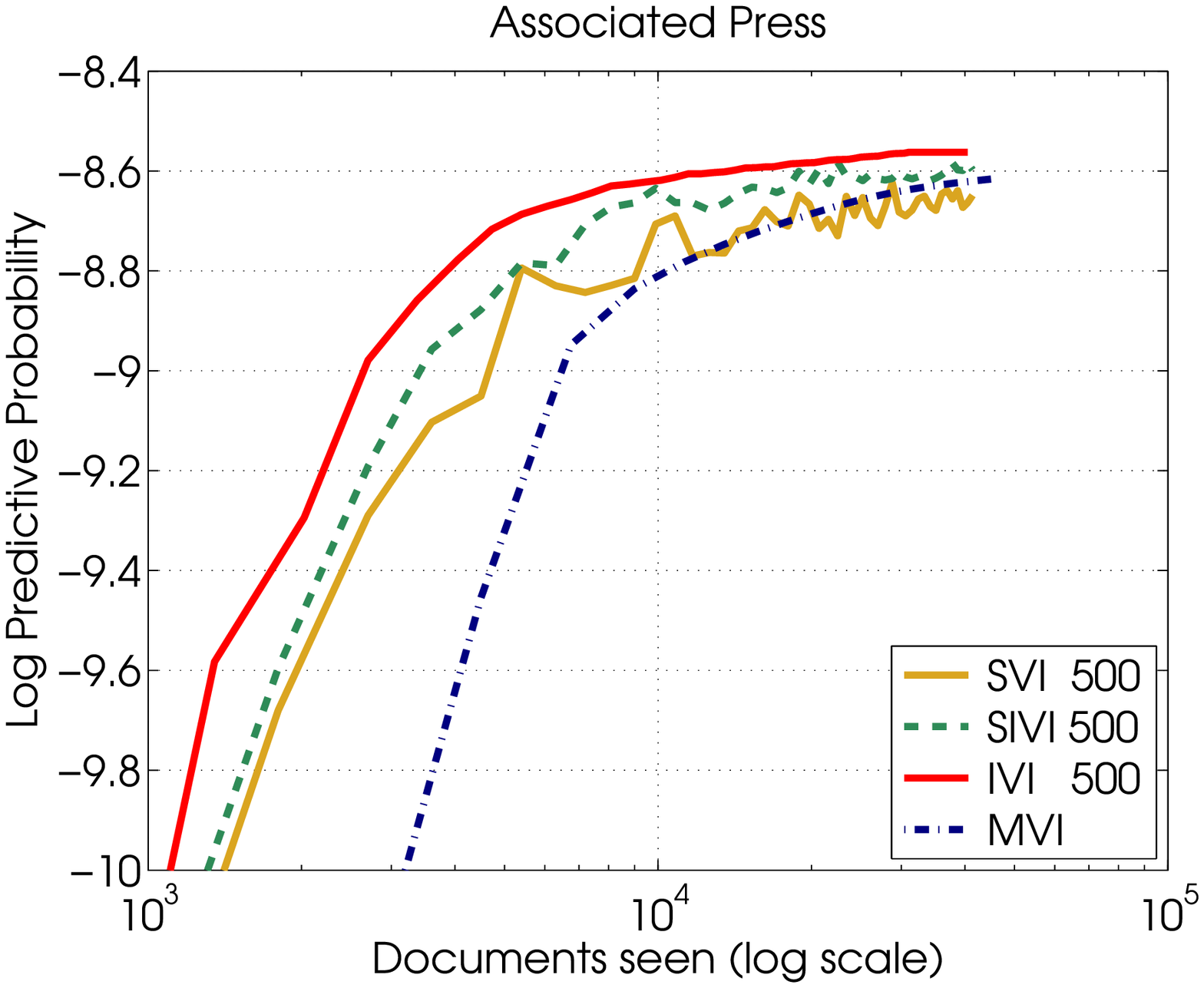}
\end{minipage}
\begin{minipage}[b]{0.24\linewidth}
\centering
\includegraphics[scale=0.23]{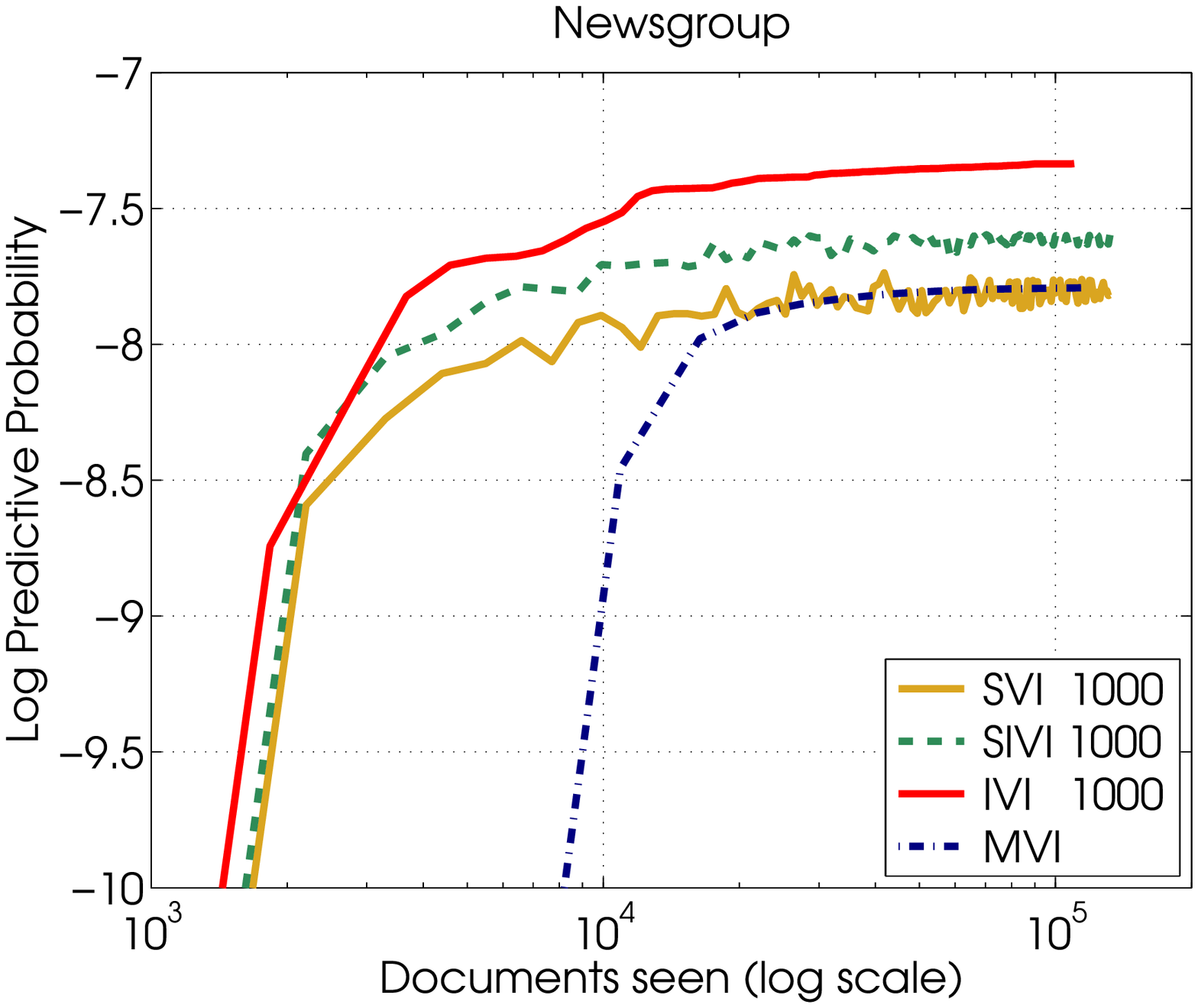}
\end{minipage}
\begin{minipage}[b]{0.24\linewidth}
\centering
\includegraphics[scale=0.23]{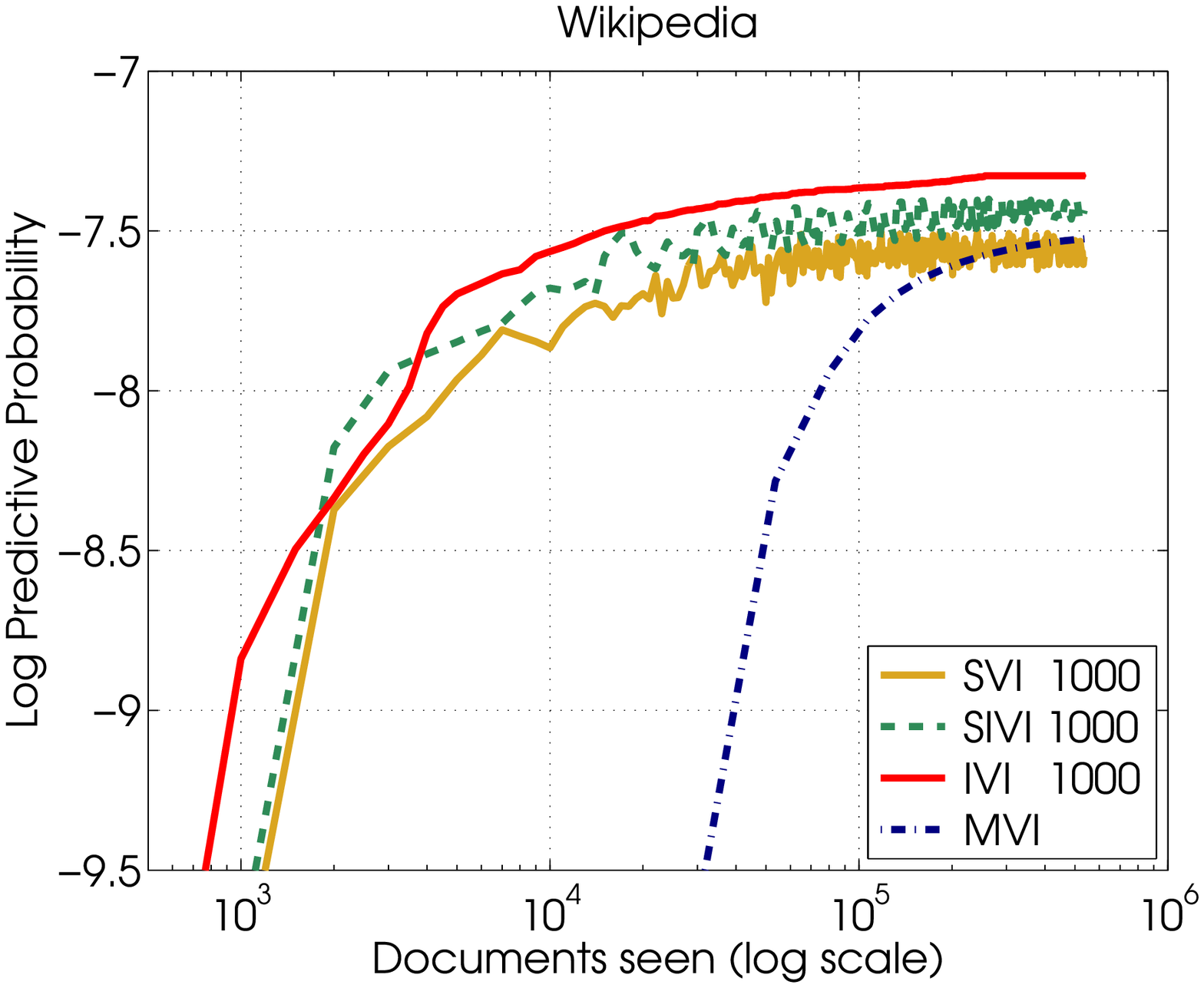} 
\end{minipage}
\begin{minipage}[b]{0.24\linewidth}
\centering
\includegraphics[scale=0.23]{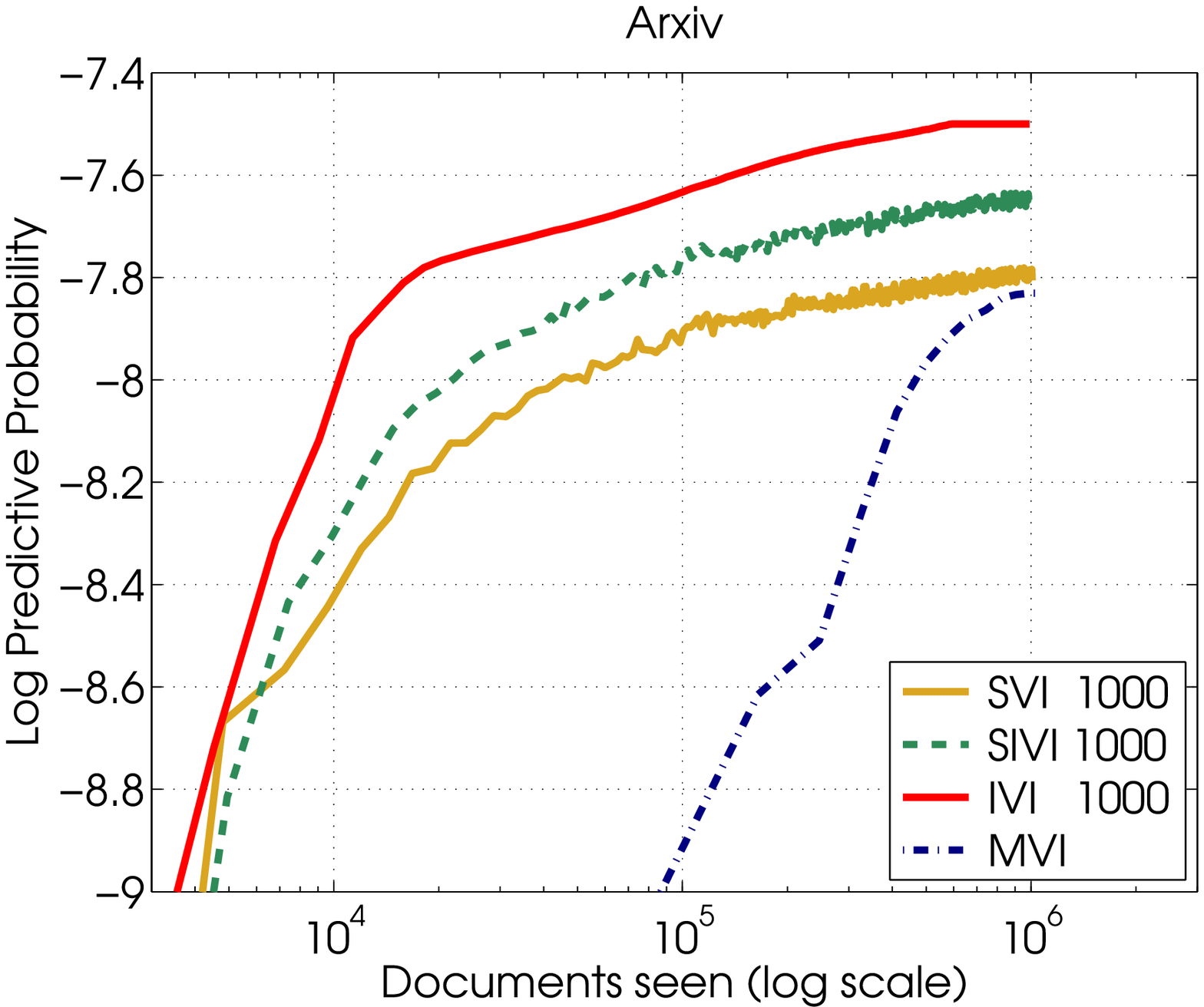} 
\end{minipage}
\caption{Per-word predictive probabilitiy for LDA  as a function of the number of processed documents. We compare results for the Associated Press, Newsgroup, Wikipedia and Arxiv data sets. Incremental approaches (IVI and S-IVI) converge to a higher value on all datasets. We reported results for 2 mini-batch sizes.}
\label{fig:compMethods}
\end{figure*}

\paragraph*{Hardware:}  Experiments were all run on a 32-core machine with 3.6 GHz Intel Core i7-3820 processors and a total of 128GB of RAM.

\paragraph*{Data:} We benchmark IVI on four corpora: Associated Press articles 
,  Newsgroup documents 
, Wikipedia articles and the scientific abstracts from Arxiv repository~\cite{mandt2014smoothed}. Besides, we used two additional large corpora to evaluate D-IVI: reviews from Amazon website and New York Times articles~\cite{mandt2014smoothed}.
The characteristics of the datasets are reported in Table~\ref{table:datasets}. 

\paragraph*{Experimental Setup:} To quantitatively evaluate the model, we estimate the predictive probability over the vocabulary~\cite{blei2003latent}. We wish to achieve high average per-word likelihood on held-out test documents. Under this metric, a higher score is better, as a better model will assign a higher probability to the held-out words. 
We learn the topics on the training corpus. We use half of each test document to estimate its topics proportions and use the remainder to compute the predictive distribution over the vocabulary. In all the experiments, we set the number of topics $K$ to 100, the Dirichlet hyperparameters $\alpha_0$ to 0.5 and $\beta_0$ to 0.05. For stochastic methods, we set the forgetting constant $\kappa$ to 0.9 and the delay $\tau$ to 1.

\subsection{IVI Prediction Results}

In the first set of experiments, we compare the different inference algorithms for LDA, using our own implementation of MVI, SVI and  IVI. 
Figure~\ref{fig:compMethods} shows that IVI converges to a solution which is comparable or better than MVI, SVI and S-IVI and IVI converges faster than the other algorithms. 
We first compare the performances of IVI and MVI at the point where MVI converges to a solution. IVI yields the same result after processing half (Newsgroup) to tenth (Arxiv) of the documents that MVI has processed. 
Besides, we observe that IVI gives consistently better predictive performance than MVI when both of them converges to a solution.

In Section~\ref{sec:ivi}, we mentioned that S-IVI does not maintain strictly accurate sufficient statistics, but it uses statistics computed as decaying average of recently visited data. Hence, it requires less memory than IVI and improves SVI in terms of accuracy and speed. Figure~\ref{fig:compMethods} provides the experimental support for these claims.

In the second set of experiments, we evaluate IVI with various mini-batch sizes by computing the average predictive log likelihood on the test set.

\begin{figure}[ht!]
\hspace*{-2mm}
\begin{minipage}[b]{0.49\linewidth}
\centering
\includegraphics[scale=0.23]{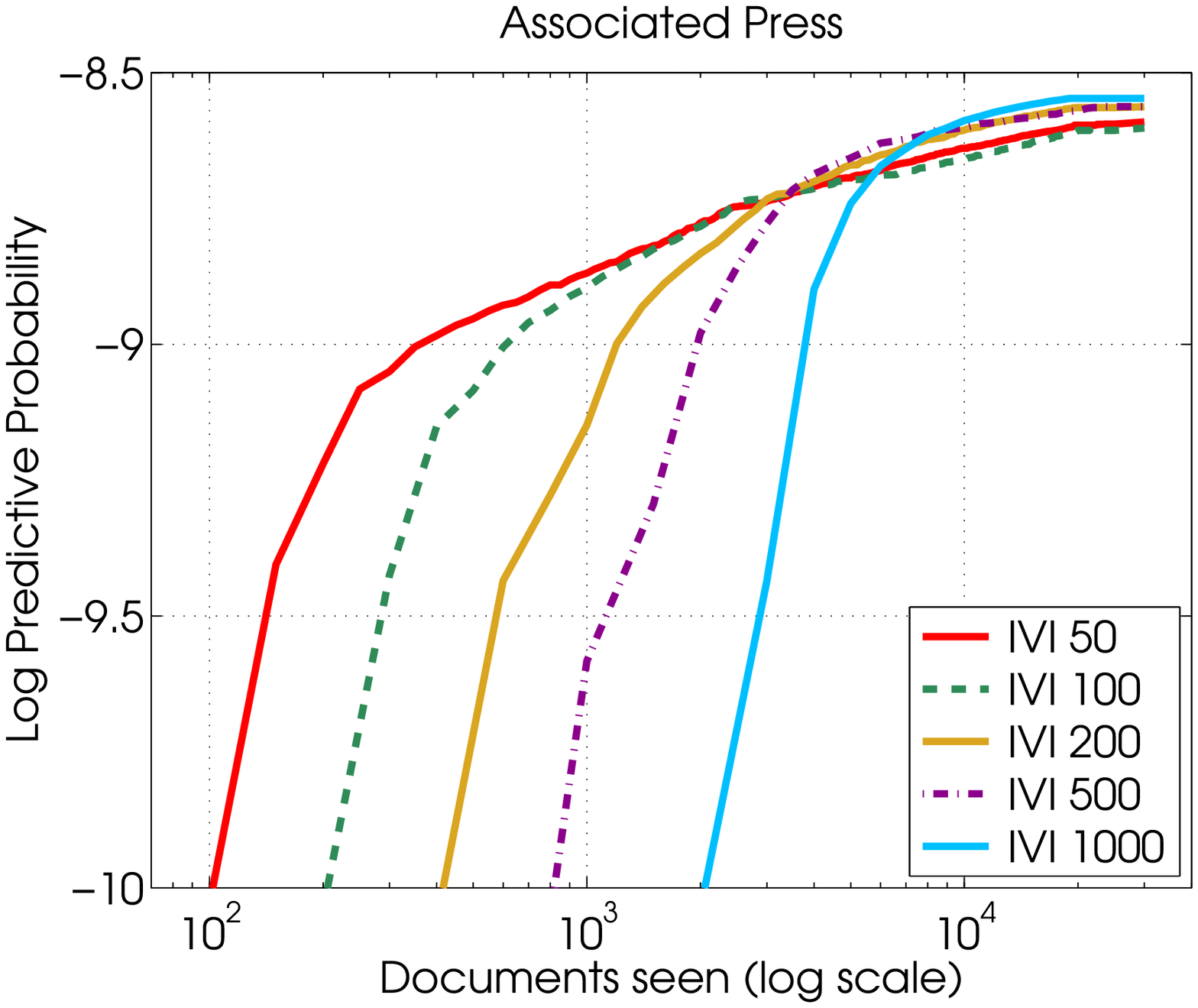}
\end{minipage}
\begin{minipage}[b]{0.49\linewidth}
\centering
\includegraphics[scale=0.23]{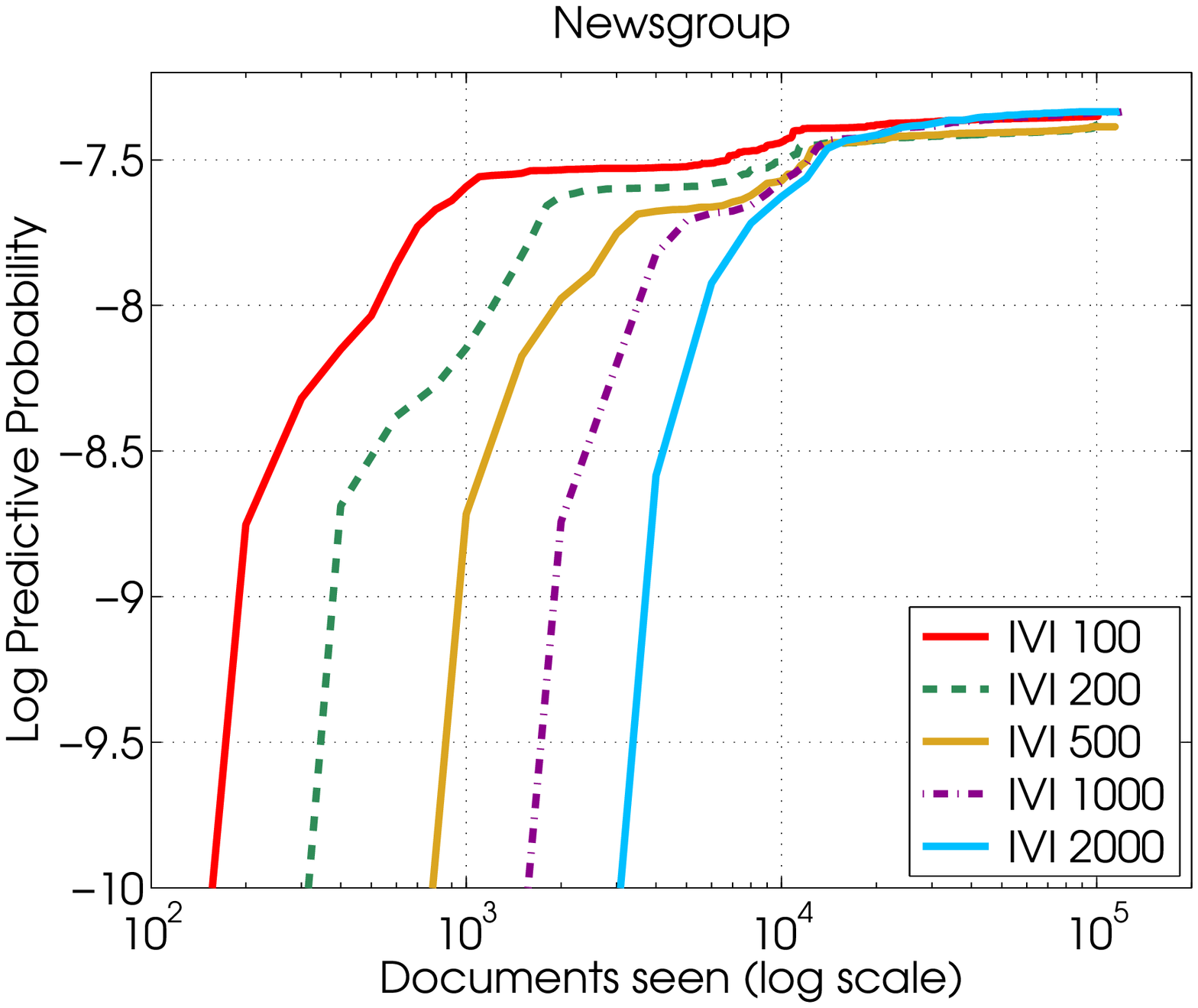}
\end{minipage}
\\
\hspace*{-2mm}
\begin{minipage}[b]{0.49\linewidth}
\centering
\includegraphics[scale=0.23]{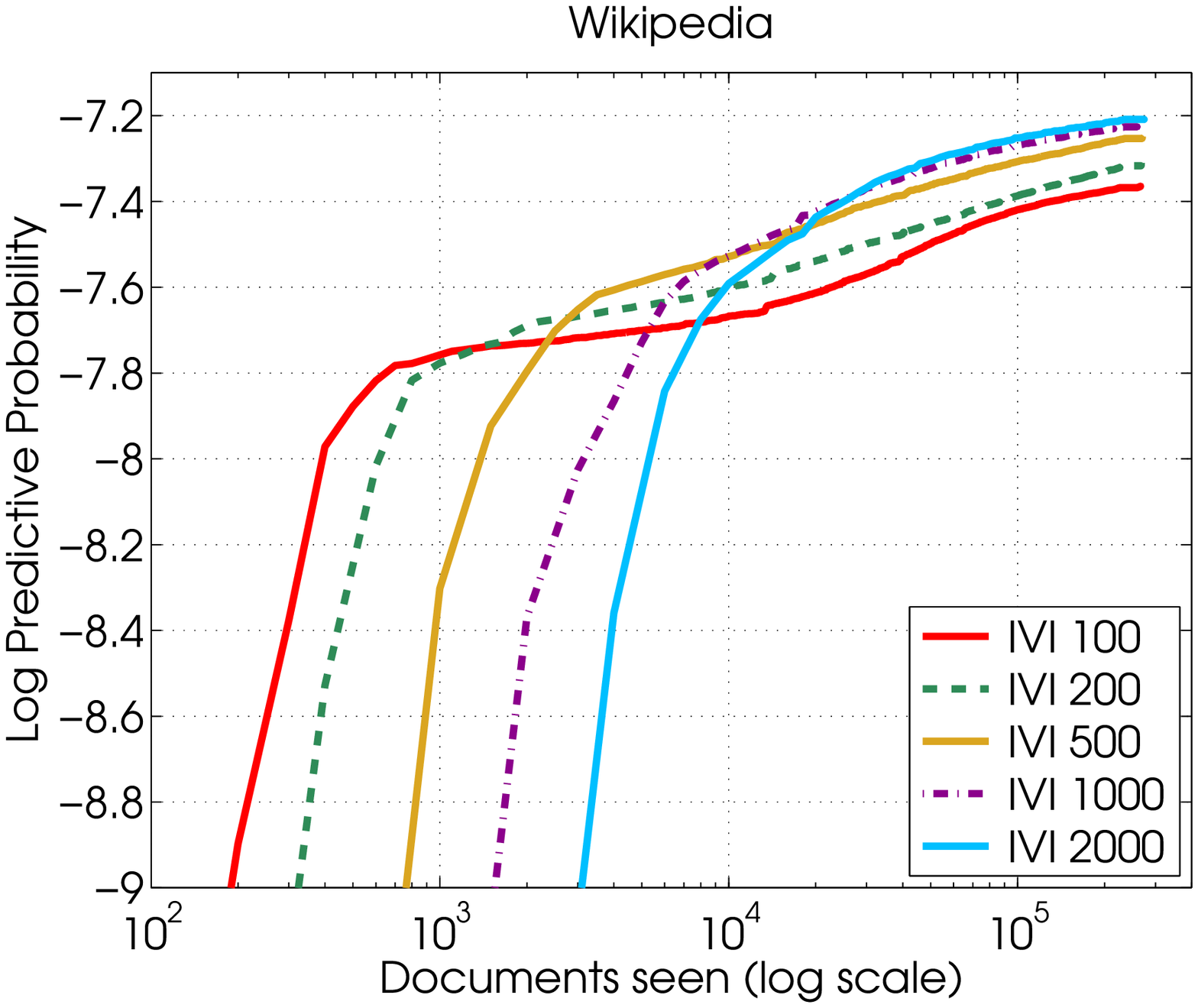}
\end{minipage}
\begin{minipage}[b]{0.49\linewidth}
\centering
\includegraphics[scale=0.23]{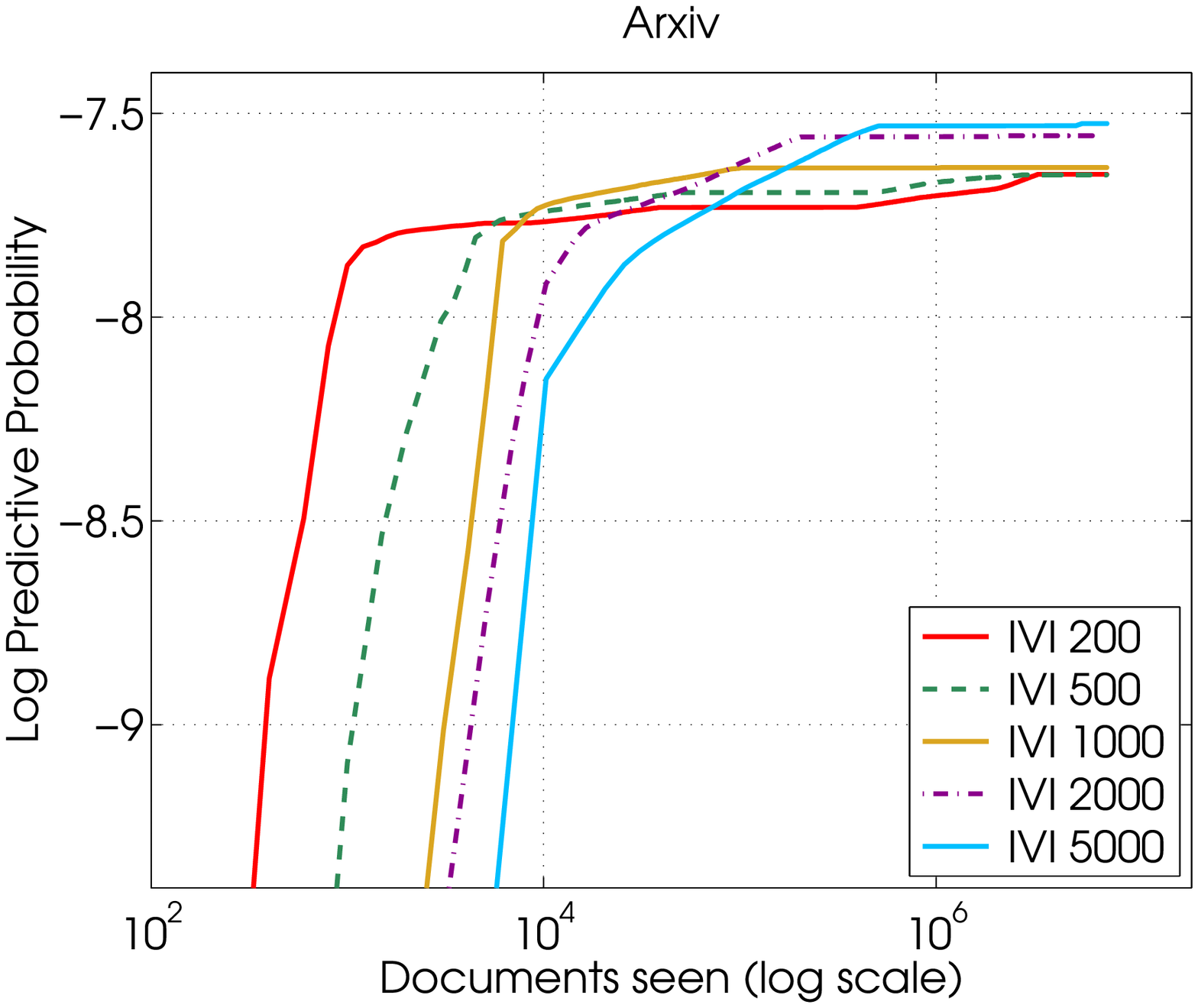}
\end{minipage}
\caption{Per-word predictive probabilitiy for LDA as a function of the number of documents. Each panel compares different values of the mini-batch size on the Associated Press, Newsgroup, Wikipedia and Arxiv data sets. IVI on the full data converges faster when a smaller batch size is used.}
\label{fig:compBatchSizeIVI}
\end{figure}

Next, we turn our attention to Figure~\ref{fig:compBatchSizeIVI}. Fixing the hyperparameters and the number of topics, we explored the effect various mini-batch sizes on all four corpora. IVI converges faster to a good solution for smaller ones. However, larger mini-batches lead to better final performance.

\subsection{D-IVI Convergence and Speed-up Results}

The purpose of these experiments is to measure speed-ups obtained with D-IVI. We report the performance of D-IVI on a single processor which corresponds to S-IVI for reference; and compare it to the performance of D-IVI for a varying number of processors. We are interested in two aspects of performance: the quality of the model learned and the time taken to learn the model. We record wall clock time and the log predictive probability on Customer Review, New York Times and Arxiv corpus. In the experiments, computations were done on $P$ processors for D-IVI where $P=\lbrace 1, 2, 4, 8, 16, 32 \rbrace$. These results are averaged over 5 runs with random initializations. 
\begin{figure}[htb!]
\hspace*{-2mm}
\begin{minipage}[b]{0.49\linewidth}
\centering
\includegraphics[scale=0.23]{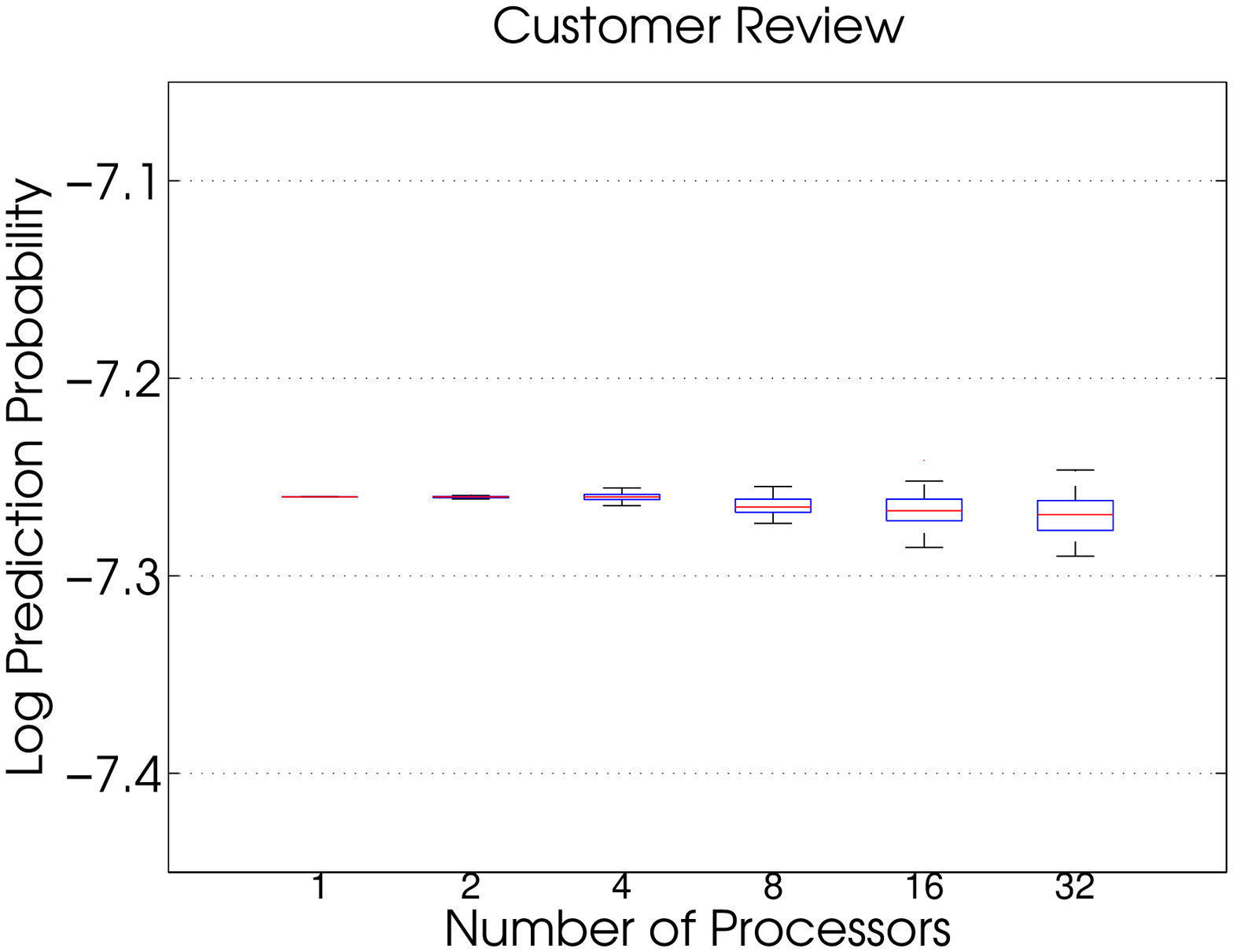}
\end{minipage}
\begin{minipage}[b]{0.49\linewidth}
\centering
\includegraphics[scale=0.23]{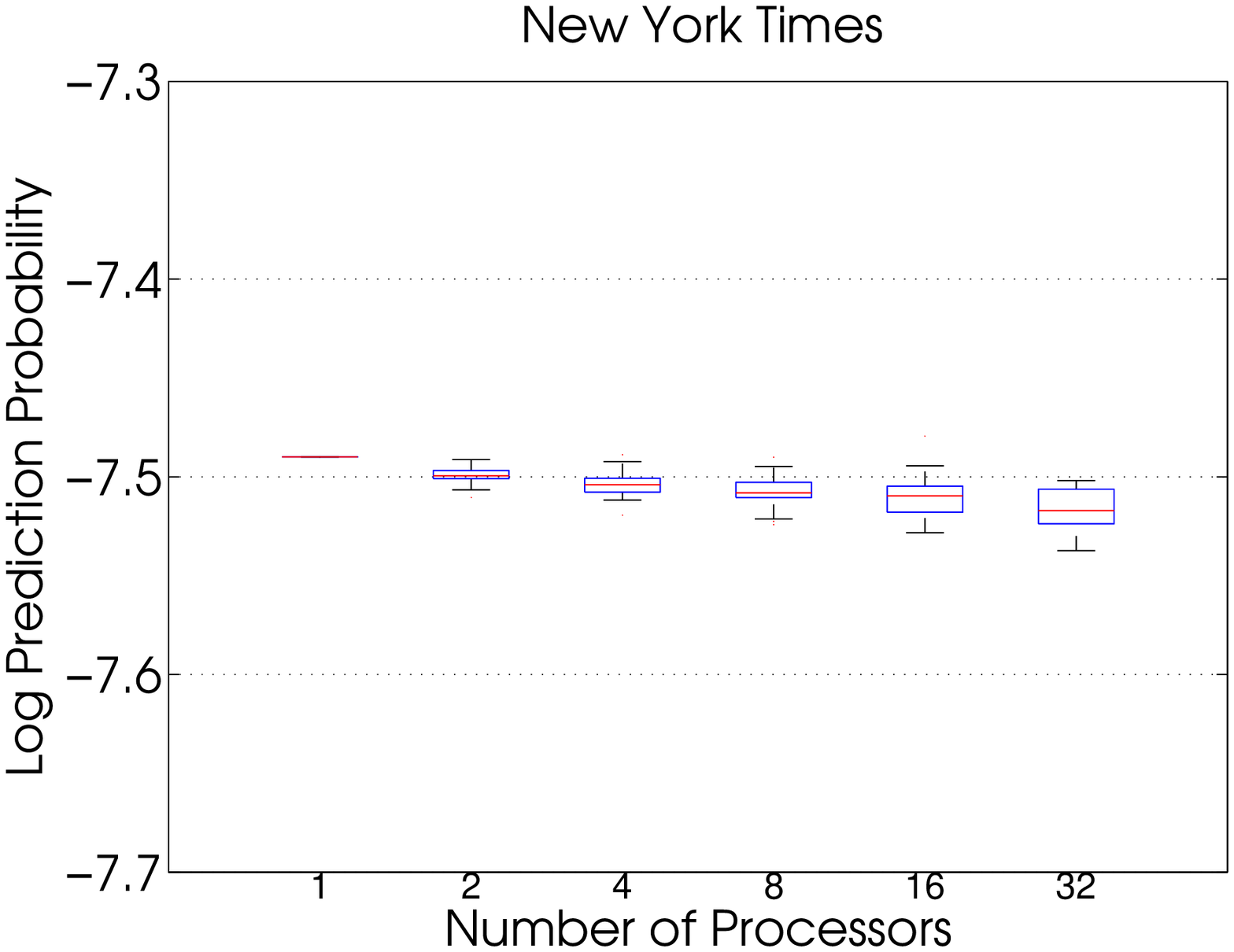} 
\end{minipage}
\\
\hspace*{-2mm}
\begin{minipage}[b]{0.49\linewidth}
\centering
\includegraphics[scale=0.23]{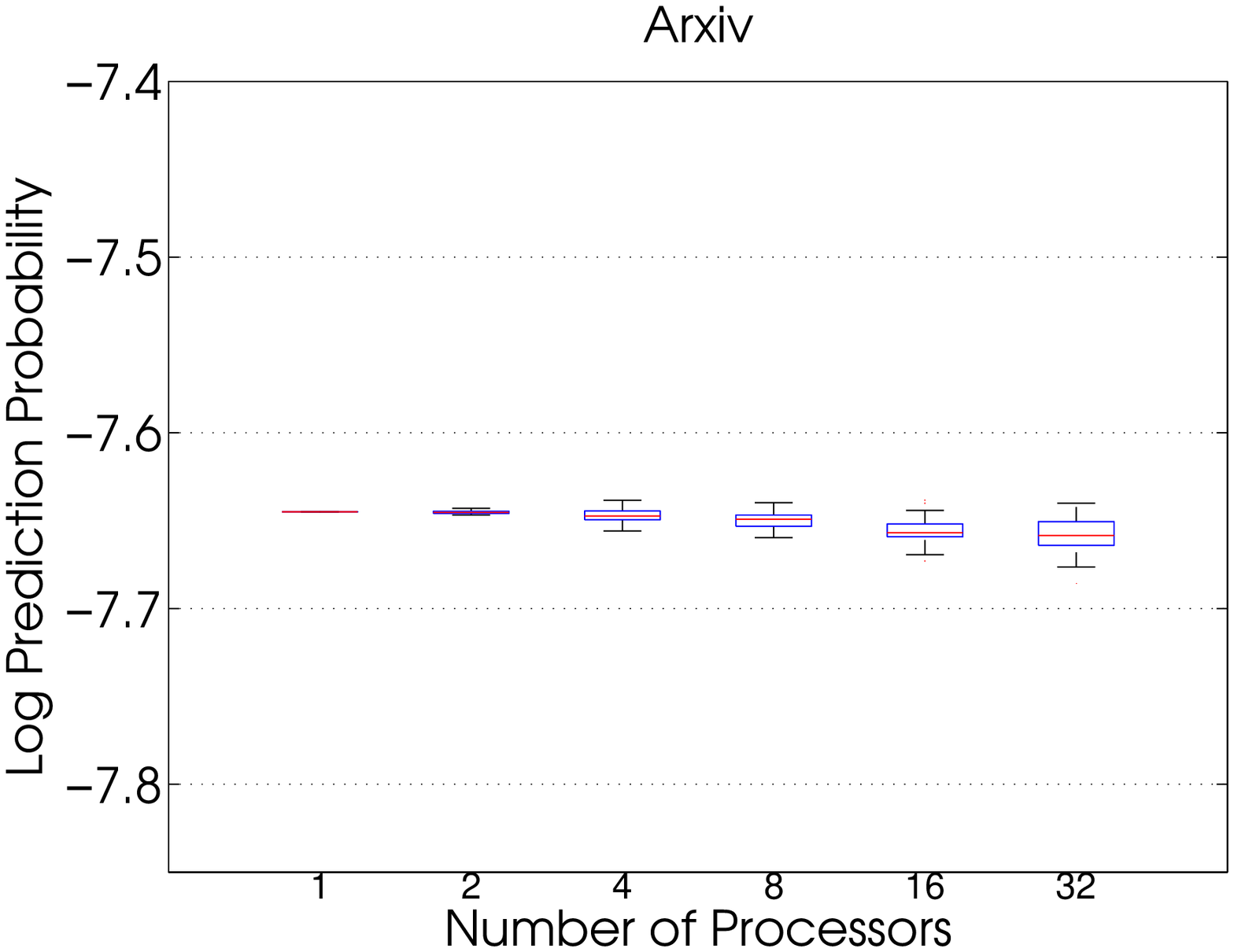}
\end{minipage}
\begin{minipage}[b]{0.49\linewidth}
\centering
\includegraphics[scale=0.23]{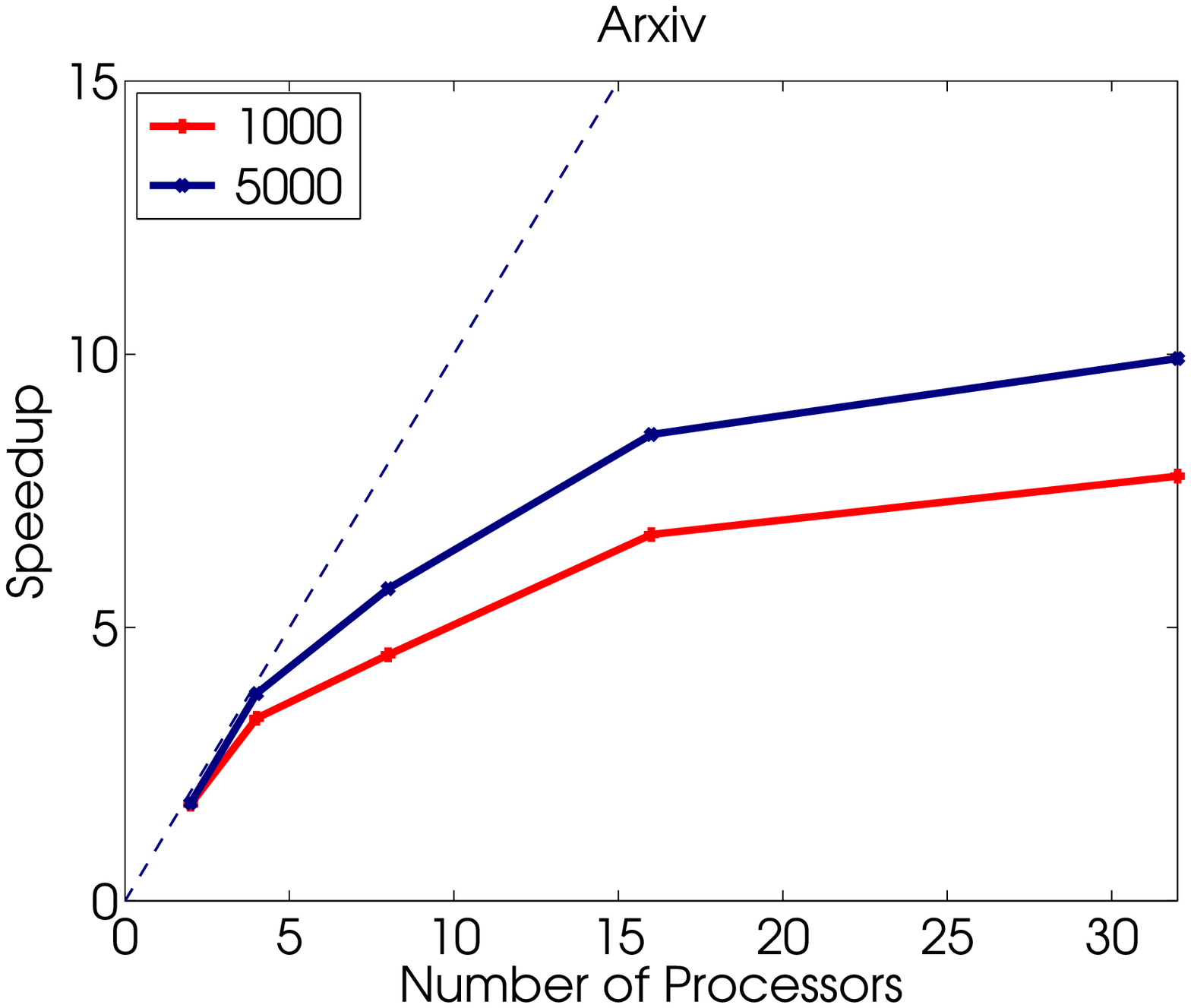} 
\end{minipage}
\caption{Log predictive probability comparisons for S-IVI and D-IVI for different number of processors on Arxiv, Customer Review and NYT. \emph{Bottom right:} Speed-up results of D-IVI for varying number of processors with respect to single processor for Arxiv. Higher speed-up is obtained with larger mini-batches due to the diminished communication overhead.}
\label{fig:compSpeedups2}
\end{figure}
\renewcommand{\arraystretch}{1.5}
\begin{table}[ht!]
\caption{Log-prediction-probability (LPP) and runtime (in terms of seconds per iteration) of the D-IVI for different number of mini-batch sizes and number of processors.}
\begin{center}
\hspace*{-2mm}
\scalebox{0.65}{
\begin{tabular}{| c | c |  c |  c  | c  | c  | c  | c  |  c  |}
\hline     &      &     &  \multicolumn{6}{c |}{Number of Processors}   \\  \cline{3-9}
Dataset  &  \pbox{20mm}{\vspace*{-5mm} Mini-batch \\ Size}   &    &  1  &  2  &  4  &  8  & 16  & 32   \\  \hline
\multirow{6}{*}{\pbox{20mm}{Customer \\ Review \\ (CR)}} & \multirow{2}{*}{1000}  &  LPP  &  -7.25  &   -7.25   &  -7.25   &  -7.28   &  -7.28     &   -7.28   \\  \cline{3-9}
&   &  Time  &  13626  &  8015    &   4367   &  3299    &   2428   &    2259    \\ \cline{2-9}
& \multirow{2}{*}{2000}  &  LPP  &  -7.26    &   -7.26   &  -7.26    &   -7.26   &  -7.28  &  -7.28   \\  \cline{3-9}
&   &  Time  &  13162  &  7607    &   4126   &  3082    &   2237   &    2113  \\ \cline{2-9}
& \multirow{2}{*}{5000}  &  LPP  &   -7.21   &   -7.21   &   -7.24   &   -7.24   &  -7.24  &  -7.24  \\ \cline{3-9}
&   &  Time  &  13043  &   7538   &   3875    &  2757    &   1883  &    1659    \\  \hline \hline
\multirow{6}{*}{\pbox{20mm}{New York \\ Times \\ (NYT)}}  &   \multirow{2}{*}{1000}  &  LPP  &  -7.49  &   -7.49   &  -7.51   &  -7.51   &  -7.51    &   -7.51  \\  \cline{3-9}
&   &  Time  &    12935    &   6916  &  3902   &   2879   &  1987     &  1728    \\  \cline{2-9}
& \multirow{2}{*}{2000}  &  LPP     &  -7.49   &  -7.49    &  -7.51     &  -7.51     &  -7.51   &  -7.51  \\ \cline{3-9}
&   &  Time  &    12906    &  6826   &  3716    &   2648   &  1956     &  1701    \\ \cline{2-9}
& \multirow{2}{*}{5000}  &  LPP    &  -7.48    &  -7.48    &  -7.48     &  -7.48   &  -7.50  &  -7.50 \\ \cline{3-9}
&   &  Time  &  12427   &    6510   &  3407    &  2360    &  1748     &  1428   \\  \hline \hline
\multirow{6}{*}{\pbox{20mm}{Arxiv}} &   \multirow{2}{*}{1000}  &  LPP  &  -7.63  &   -7.63   &  -7.63   &  -7.65   &  -7.66    &   -7.66  \\  \cline{3-9}
&   &  Time   &   16996  &   9601   &   5087    &   3776   &    2534   &  2185  \\ \cline{2-9}
& \multirow{2}{*}{2000}  &   LPP  &  -7.55     &  -7.55     &  -7.56     &  -7.56   &  -7.56  &  -7.56  \\ \cline{3-9}
&   &  Time   &  17845   &   9857   &  5110    &   3678    &   2453    &  2158   \\  \cline{2-9}
& \multirow{2}{*}{5000}  &   LPP  &  -7.52     &  -7.52     &  -7.52     &  -7.54   &  -7.54  &  -7.54  \\ \cline{3-9}
&   &  Time   &  17957   &  10030  &   4760    &   3228   &   2105    &  1835    \\ \hline
\end{tabular}}
\label{table:multiThread}
\end{center}
\end{table}
The results in Table~\ref{table:multiThread} and Figure~\ref{fig:compSpeedups2} show that the log predictive probability is essentially the same for the distributed models as their single-processor versions at $P=1$. Errors due to the stale parameters caused slight variations in performance of D-IVI. This variation increases with the number of processors. This is shown in Figure~\ref{fig:compSpeedups2}, where we report the box and whiskers plot of the log predictive probability.  
\begin{figure*}[htb!]
\begin{minipage}[b]{0.34\linewidth}
\centering
\includegraphics[scale=0.22]{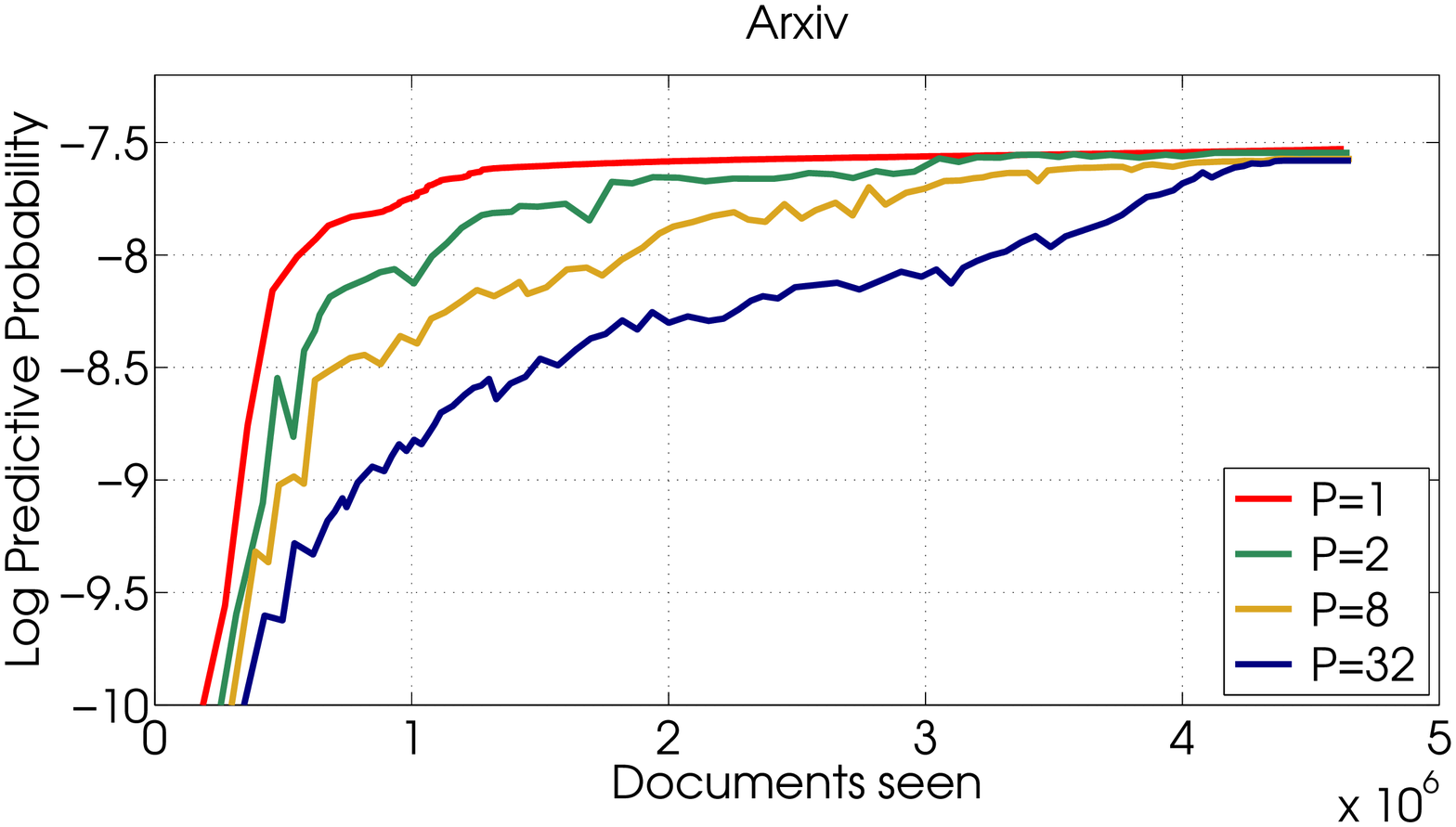}
\end{minipage}
\begin{minipage}[b]{0.33\linewidth}
\centering
\includegraphics[scale=0.22]{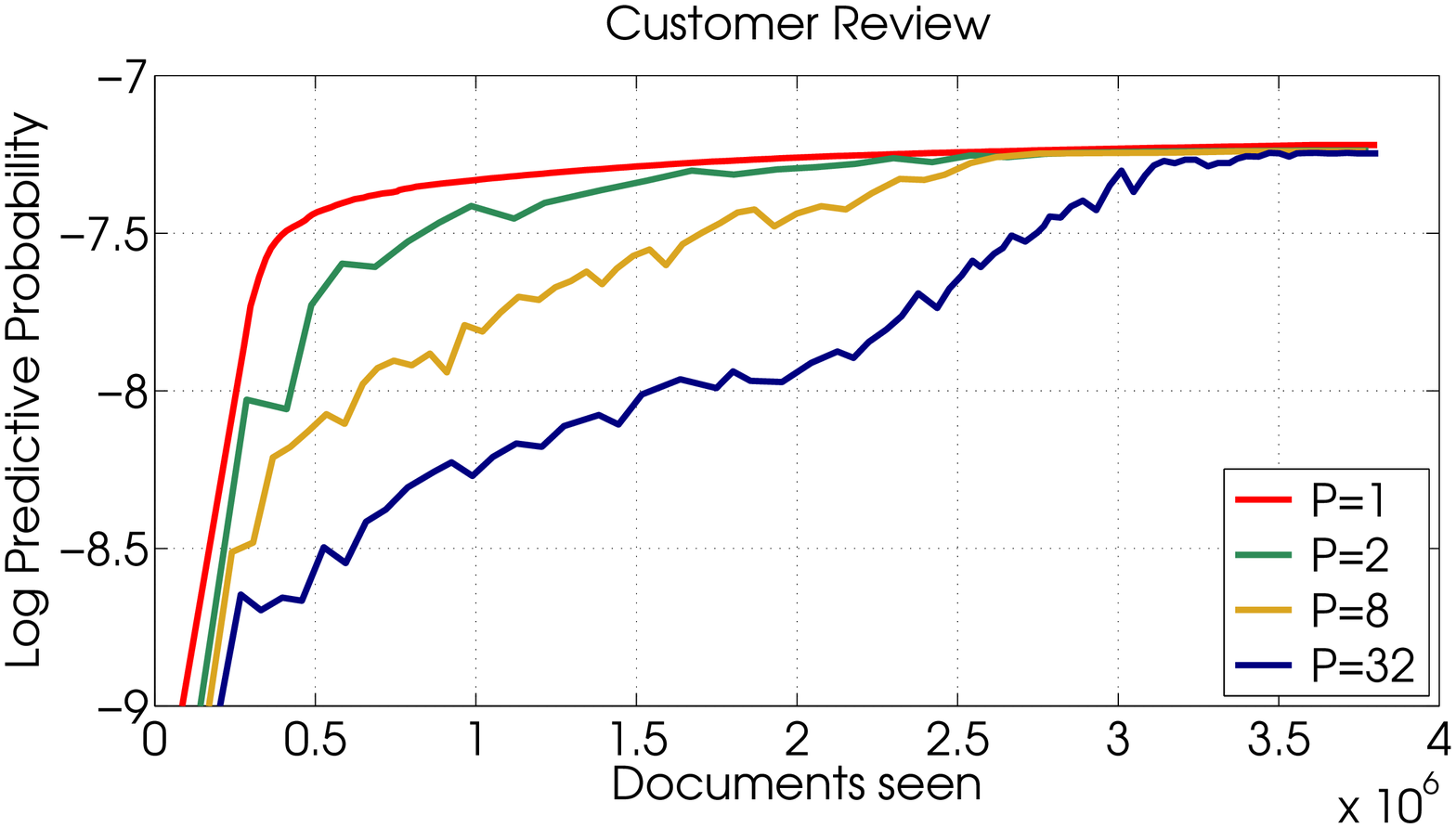}
\end{minipage}
\begin{minipage}[b]{0.33\linewidth}
\centering
\includegraphics[scale=0.22]{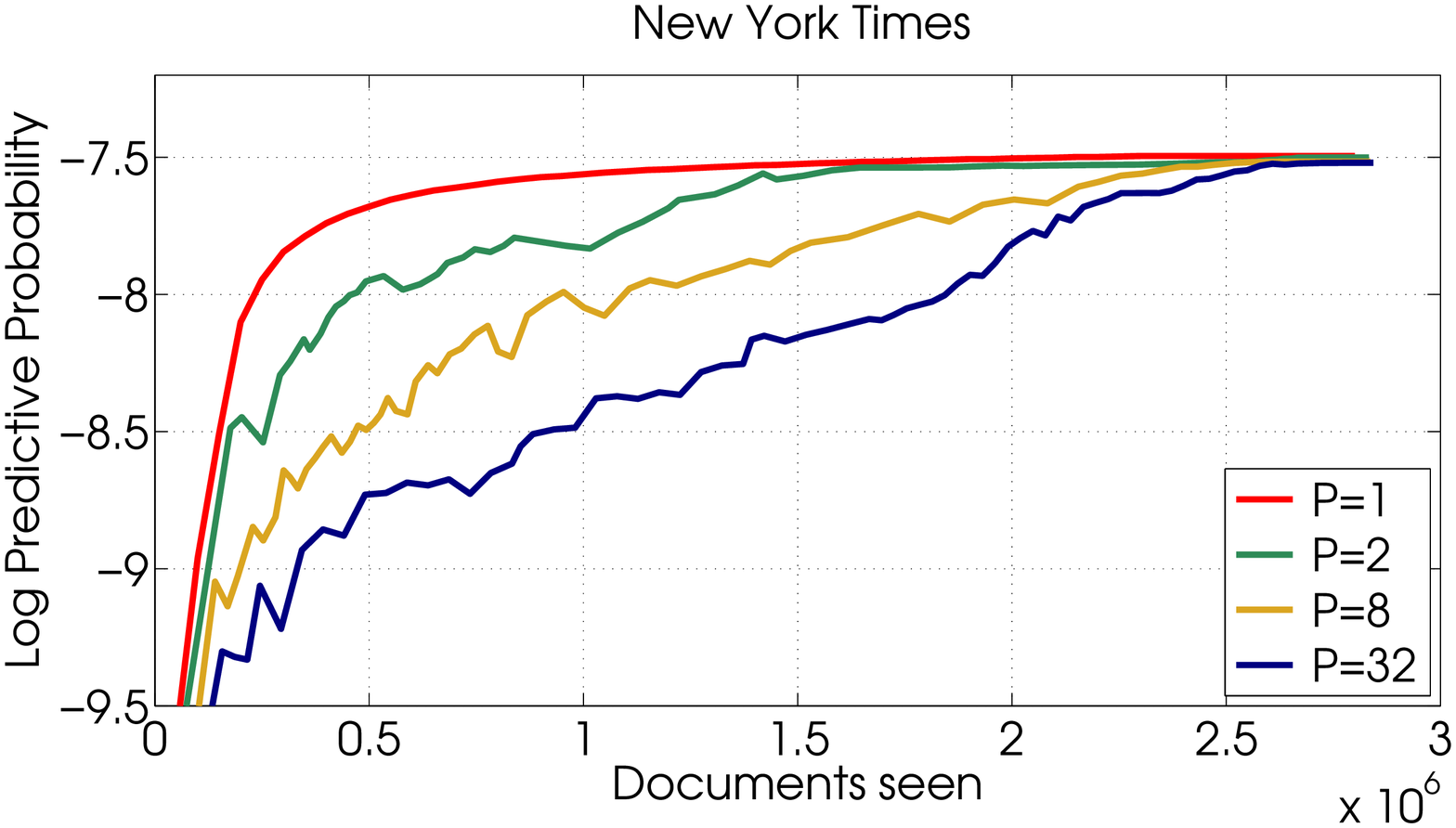} 
\end{minipage}
\caption{Convergence results (per-word predictive probabilitiy for LDA model as a function of number of documents processed so far) for D-IVI on Arxiv, Customer Review and New York Times for varying number of processors. As the number of processors increases, the rate of convergence slows down.}
\label{fig:compSpeedups}
\end{figure*}

One of the main motivation for developing D-IVI is to reduce computation time while retaining performance. The speed-up  results shown in Table~\ref{table:multiThread} and Figure~\ref{fig:compSpeedups2} (bottom-right) demonstrate that the improvement in convergence speed by increasing the number of processors is to be mitigated by the communication overhead. When number of processors is large, the data subset assigned to each processor gets smaller. In this case, each  update is less informative, and more iterations are needed for convergence. To overcome communication overhead, we have used larger mini-batch size. Hence, more information are collected in each global parameter update, and so the number of iterations required for convergence is reduced.

D-IVI increases inference speed. We observe a $\sim$1.8 times speed-up for all three corpora when using $P$=2 processors; and $\sim$7.8, $\sim$8.6 and $\sim$9.9 times speed-up respectively for CR, NYT and Arxiv datasets when using $P$=32 processors. These results suggest that asynchronous D-IVI converges to a solutions that exhibit a performance close to one obtained with S-IVI. 

\paragraph*{Simulated Delays:} Next, we add delays to some workers to explore the robustness of D-IVI. Figure~\ref{fig:compSpeedups} provides results when each processor sleeps with $0.5$ probability for a small amount of time before sending the latest sufficient statistics correction to the master. The delay length is chosen randomly from a normal distribution with the mean $\mu$ (in seconds) and $\sigma = \mu/5$. We have specified the upper limit of $\mu$ as twice the average time required to compute the sufficient statistics of a mini-batch.

Here, we report performance by plotting log predictive probability against number of documents seen so far.
Figure~\ref{fig:compSpeedups} shows that as the number of processors increases, the rate of convergence slows down, since more iterations are needed for information to propagate to all the processors. However, it is important to note that one iteration in real time of D-IVI is up to number of processors times faster than one iteration of S-IVI, so D-IVI converges much more quickly than S-IVI (see Table~\ref{table:multiThread} for time results).

\begin{figure}[ht!]
\centering
\includegraphics[scale=0.3]{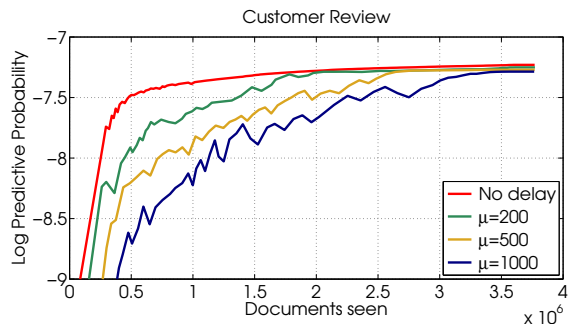}
\caption{Convergence of D-IVI when a delay is encountered. The delay time are sampled from $\mathcal{N}(\mu, \sigma^2)$, for several values for $\mu$. As the $\mu$ increases, the rate of convergence slows down. While curves with $\mu$= 500 and 1000 appear less smooth than the others, they are still heading steadily toward convergence. As the number of processors increases, the rate of convergence slows down}
\label{fig:artDelay}
\end{figure} 

Finally, we test if D-IVI is robust to extremely stale parameters by increasing the delay. Figure~\ref{fig:artDelay} shows the results of this case. For CR corpora, the computation time of the sufficient statistics for a mini-batch of 1000 documents is ~26 seconds in average. Here, each processor sleeps with $0.25$ probability and the average delay is set to twice (50 seconds, $\mu$=200), 5-times and 10-times the computation time for a mini-batch. 

We see that, the D-IVI algorithm still converges even with considerable delays of 5 and 10 times the processing time for a mini-batch. Despite no formal convergence guarantees, D-IVI algorithm performs well empirically in all experiments we conducted on the three real-world data sets considered. 


\section{Conclusion}

We introduced incremental variational inference as an alternative to stochastic variational inference. The algorithm does not require to adjust the learning rate. We showed experimentally that the incremental approach converges faster and often to a better local optimum of the variational objective. Incremental variational inference processes documents sequentially. It scales thus similarly to stochastic variational inference and is suitable when we can afford to incur an additional memory cost (which scales as $O(KN)$).

We further modified incremental variational inference to accommodate a stochastic variant, which can be adapted to distributed environments. This enabled us to further scale variational inference. We showed experimentally that the proposed asynchronous algorithm is robust to noise and outdated parameters, and produces solutions that are very close to the single host solutions. The horizontal speed-up saturates when then number of processors increases as communication cost increases and more passes over the data are necessary to ensure convergence to the same level of accuracy.

We left the convergence analysis of incremental variational inference to future work, as well as its application to other probabilistic models. Indeed, the incremental variational algorithms proposed in the paper are generic. They can be applied to any model with local and global variables and are by no means restricted to their application to LDA.

\bibliography{paper_arxiv}
\bibliographystyle{icml2015}

\end{document}